\documentclass[sigconf]{acmart}
\AtBeginDocument{%
  }

\copyrightyear{2026}
\acmYear{2026}
\setcopyright{cc}
\setcctype{by}
\acmConference[KDD '26]{Proceedings of the 32nd ACM SIGKDD Conference on Knowledge Discovery and Data Mining V.2}{August 09--13, 2026}{Jeju Island, Republic of Korea}
\acmBooktitle{Proceedings of the 32nd ACM SIGKDD Conference on Knowledge Discovery and Data Mining V.2 (KDD '26), August 09--13, 2026, Jeju Island, Republic of Korea}
\acmDOI{10.1145/3770855.3818951}
\acmISBN{979-8-4007-2259-2/2026/08}

\usepackage{amsmath,amsfonts,bm}

\def\eqref#1{equation~\ref{#1}}

\def\1{\bm{1}}

\def\mX{{\bm{X}}}
\def\mY{{\bm{Y}}}

\DeclareMathAlphabet{\mathsfit}{\encodingdefault}{\sfdefault}{m}{sl}
\SetMathAlphabet{\mathsfit}{bold}{\encodingdefault}{\sfdefault}{bx}{n}

\usepackage{hyperref}
\usepackage{url}
\usepackage{graphicx}
\usepackage{booktabs}
\usepackage{microtype}
\usepackage[most]{tcolorbox}
\usepackage{multirow}
\usepackage[table,xcdraw,usenames,dvipsnames]{xcolor}
\usepackage{enumitem}
\usepackage{float}
\usepackage{caption}
\usepackage{subcaption}
\usepackage{comment}

\newcommand{\dataset}{SF$^2$Bench}
\begin{document}

\title{ Uncovering Insights of Compound Flooding with Data-Driven AI }

\author{Xu Zheng}
\orcid{0000-0001-8442-0691}
\affiliation{%
  \institution{Florida International University}
  \city{Miami}
  \state{FL}
  \country{USA}
}
\email{xzhen019@fiu.edu}

\author{Chaohao Lin}
\orcid{0000-0001-9737-8007}
\affiliation{%
  \institution{Florida International University}
  \city{Miami}
  \state{FL}
  \country{USA}
}
\email{clin027@fiu.edu}

\author{Sipeng Chen}
\orcid{0009-0008-5576-6499}
\affiliation{%
  \institution{Florida State University}
  \city{Tallahassee}
  \state{FL}
  \country{USA}
}
\email{sc25bg@fsu.edu}

\author{Zhuomin Chen}
\orcid{0009-0002-4406-7947}
\affiliation{%
  \institution{Florida International University}
  \city{Miami}
  \state{FL}
  \country{USA}
}
\email{zchen051@fiu.edu}

\author{Jimeng Shi}
\orcid{0009-0002-7268-0431}
\affiliation{%
  \institution{UIUC}
  \city{Champaign}
  \state{IL}
  \country{USA}
}
\email{jimeng8@illinois.edu}

\author{Jayantha Obeysekera}
\orcid{0000-0002-7038-1668}
\affiliation{%
  \institution{Florida International University}
  \city{Miami}
  \state{FL}
  \country{USA}
}
\email{jobeysek@fiu.edu}

\author{Jingchao Ni}
\orcid{0000-0002-2986-6612}
\affiliation{%
  \institution{University of Houston}
  \city{Houston}
  \state{TX}
  \country{USA}
}
\email{jni7@uh.edu}

\author{Wei Cheng}
\orcid{0000-0001-5456-626X}
\affiliation{%
  \institution{NEC Lab America}
  \city{Princeton}
  \state{NJ}
  \country{USA}
  }
\email{weicheng@nec-labs.com}

\author{Jason Liu}
\orcid{0000-0001-8222-4013}
\affiliation{%
  \institution{Florida International University}
  \city{Miami}
  \state{FL}
  \country{USA}
}
\email{liux@fiu.edu}

\author{Dongsheng Luo}
\orcid{0000-0003-4192-0826}
\authornote{Corresponding author. This work was primarily conducted while he was at Florida International University.}
\affiliation{%
  \institution{Singapore Management University}
  \country{Singapore}
}
\email{luodongsheng01@gmail.com}

\renewcommand{\shortauthors}{Xu Zheng et al.}

\begin{abstract}
Compound flooding, driven by nonlinear interactions between multiple hydrometeorological factors, poses a significant challenge to hazard prevention.  Existing forecasting approaches, whether physics-based or data-driven, often emphasize temporal patterns while underexploring how multiple interacting factors jointly shape flood dynamics. To address this problem, we conduct a large-scale data-driven analysis of compound flooding in South Florida, a typical area for compound flooding, by integrating tidal conditions, rainfall, groundwater stage, and human water management activities.
Our analysis reveals three key findings: (i) models that capture temporal dynamics alone fail to represent multi-factor interactions during compound events; (ii) subsurface saturation, as reflected by groundwater levels, emerges as a dominant predictor of flood severity, often outweighing immediate rainfall intensity in this porous coastal region; and (iii) the spatial state of surrounding monitoring stations within a finite effective radius provides critical causal context for flooding, while extending temporal history yields diminishing returns during extreme events. 
These findings suggest that compound flooding is governed more by spatially coupled system states than by long-term temporal dependencies, challenging rain-centric and sequence-dominated forecasting paradigms. By framing data-driven models as tools for scientific inquiry rather than prediction alone, this study offers new insights into the mechanisms of compound flooding and informs the design of more physically grounded early-warning systems for coastal environments. Our dataset and code are publicly available at \url{https://github.com/AslanDing/SFBench}.
\end{abstract}



\maketitle

\begin{figure}[t]
    \centering
    \includegraphics[width=0.47\textwidth]{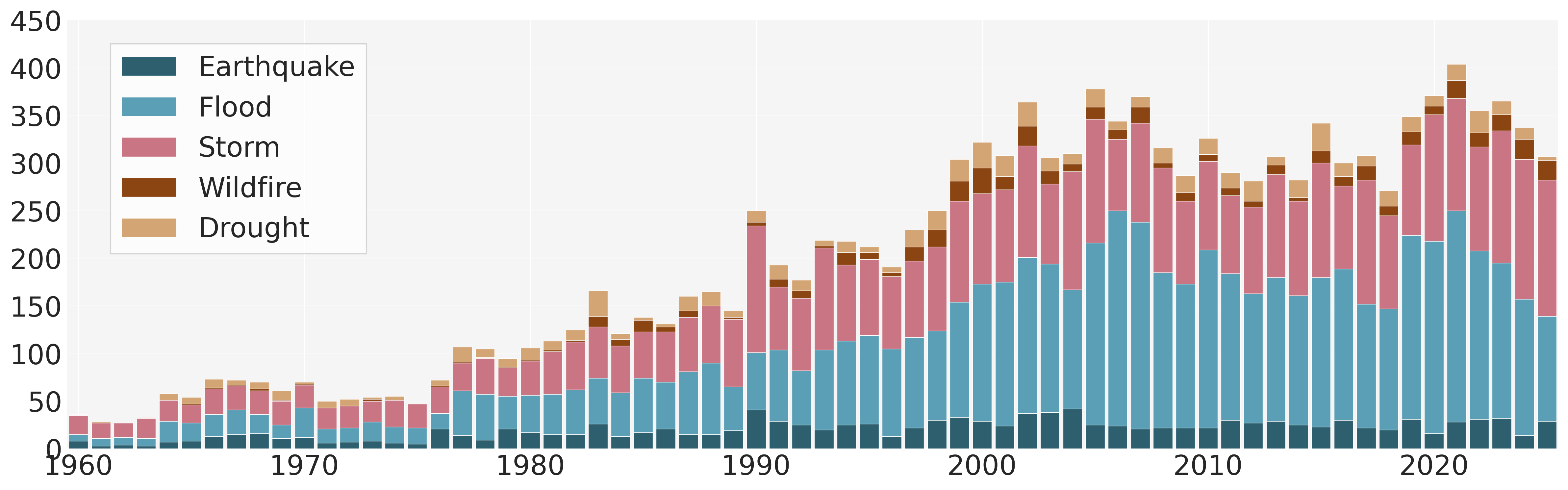}
    \vspace{-0.3cm} 
    \caption{Global historical statistics for 5 types of natural disaster events. Data source: EM-DAT~\cite{delforge2025dat}. Flood is the most common disaster. }
    \label{fig:floods}
    \vspace{-0.4cm}
\end{figure}

\section{Introduction}
Floods are among the most frequent and destructive natural hazards, causing severe environmental damage~\citep{yin2023flash}, catastrophic loss of life~\citep{jonkman2008loss}, and economic disruption~\citep{brody2007rising}, as shown in Figure~\ref{fig:floods}.  Compared with single-driver flood events, such as stream overflows or flash floods~\citep{green2024comprehensive}, compound floods arise from the concurrence of multiple interacting drivers across meteorological, hydrological, and oceanographic processes~\citep{SEBASTIAN202277}. According to previous research~\cite{bevacqua2020more}, compound flooding is projected to increase globally by over 25\% by 2100 relative to present under a high-emissions scenario.
These events are particularly difficult to predict, as nonlinear interactions among drivers can amplify flood severity beyond what any single factor would suggest. Recent studies indicate that both the frequency and spatial extent of compound floods are increasing under climate change~\citep{wahl2015increasing,wing2022inequitable,hirabayashi2013global}, underscoring the urgency of understanding their underlying mechanisms.

Despite their importance, the causal structure of compound flooding remains poorly understood. 
Classical physics-based methods simulate flood dynamics by solving partial differential equations that encode hydrodynamic processes~\citep{paniconi2015physically,yin2023physic}, such as the Hydrologic Engineering Center’s River Analysis System (HEC-RAS)~\citep{brunner1997hec}. Despite their accuracy and explainability, their application is constrained by extensive data requirements, including high-resolution terrain data, canal networks, and river geometries~\citep{sampson2015high,zang2021improving}. As a result, physics-based models are often difficult to deploy consistently across regions, limiting their utility for large-scale comparative analysis of compound flood drivers.    
In parallel, data-driven approaches based on machine learning have emerged as an alternative for flood forecasting, leveraging advances in deep learning to model complex, nonlinear relationships directly from observations. Prior studies have applied Convolutional Neural Networks (CNN)~\citep{lecun1998gradient}, Recurrent Neural Networks (RNN)~\citep{hochreiter1997long}, Graph Neural Networks (GNN)~\citep{kipf2017semisupervised}, and Transformers~\citep{vaswani2017attention} to capture temporal and spatial dependencies in flood processes. However, most existing models emphasize a single driving force and treat the compound flooding forecasting task as a temporal analysis task, overlooking the interaction of various factors. For example, Google researchers use Long Short-Term Memory networks (LSTMs)~\citep{hochreiter1997long} to forecast extreme floods with watershed streamflow data~\cite{nearing2024global}. In \cite{hess-22-6005-2018}, the authors build a model to capture the relationship between rainfall and discharge of catchments.  These methods lead to oversimplified scientific insights, compressing the complex interplay among multiple driving factors into a sequential dependency.

A fundamental obstacle to resolving these questions is the lack of integrated observational frameworks that support systematic, multi-factor analysis of compound flooding. Existing datasets~\citep{kabir2020deep,RUMA2023100951,shi2024fidlar,klingler2021lamah} are often limited in geographic scope or omit key drivers such as groundwater dynamics and human water management operations. Consequently, previous studies have largely examined compound floods through isolated case studies or restricted regional settings~\citep{xu2023impact}, making it difficult to unravel the relative importance of interacting factors or to assess how spatial context influences the causality of floods.

\begin{figure}[t]
    \centering
    \includegraphics[width=0.45\textwidth]{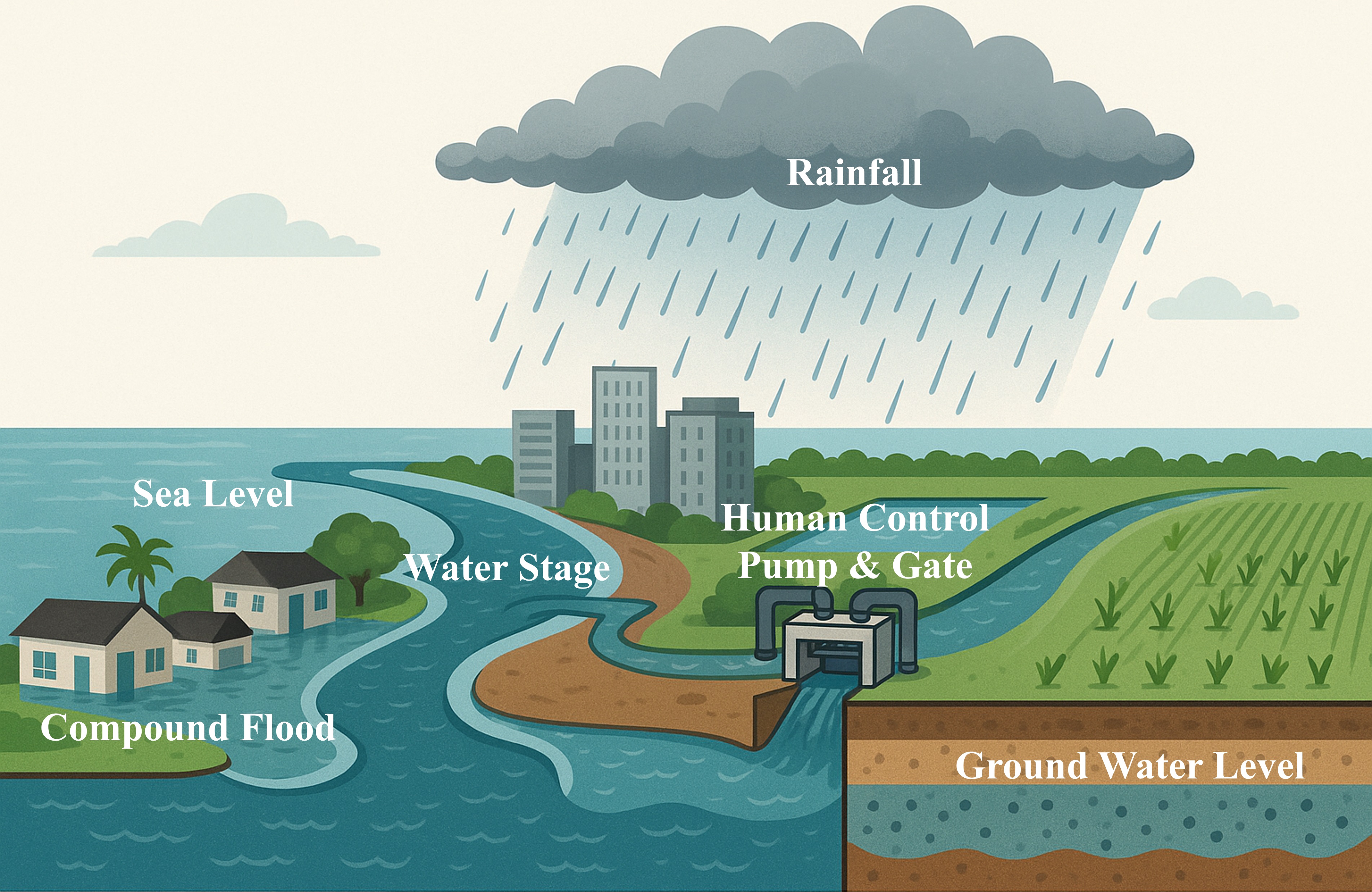}
    \vspace{-0.3cm}
    \caption{Schematic diagram of multiple factors.}
    \label{fig:sketch}
    \vspace{-0.6cm}
\end{figure}

In this study, to bridge the gap between existing methods and real-world scenarios, we leverage data-driven models not merely as forecasting tools but as analytical probes to investigate the mechanisms underlying compound flooding. We select South Florida as a typical area for the analysis, a low-lying coastal region that serves as a natural laboratory for compound flood analysis due to its porous geology, dense water management infrastructure, and frequent exposure to hurricanes and extreme precipitation~\citep{nhess-20-2681-2020}. Flooding in this region is shaped by the interaction of multiple drivers, including rainfall, tide, groundwater saturation, and human-controlled hydraulic structures, providing a uniquely rich setting for studying multi-factor flood dynamics, as shown in Figure~\ref{fig:sketch}.  To enable this analysis, we assemble a comprehensive observational dataset spanning 1985 to 2024, integrating water stage, tide, subsurface water(groundwater) stage, rainfall, and operational data from gates and pumps across 2,452 monitoring stations. Rather than treating this dataset as a benchmark alone, we use it to support controlled analyses of how different modeling paradigms respond to variations in temporal history, spatial context, and driver availability.

With a diverse set of learning architectures, including Multilayer Perceptrons (MLPs), RNNs, CNNs, GNNs, Transformers, and Large Language Models (LLM)-based approaches, we systematically examine model behavior under feature ablation, spatial perturbation, and extreme-event evaluation. The analysis yields three central findings: (i) models that depend on temporal patterns struggle to represent critical multi-factor interactions during compound flood events; (ii) subsurface saturation consistently emerges as a dominant driver of flood severity in South Florida, frequently exceeding the influence of short-term rainfall; and (iii) spatially localized system states within a finite effective radius provide essential causal context for flooding, while extending temporal history offers limited benefit during extremes.

While demonstrated in South Florida, these results reveal fundamental insights that compound flood dynamics emerge not from isolated temporal sequences but from the spatial entanglement of multi-hazard drivers. This work reframes data-driven models as generalizable tools for scientific discovery, offering a universal blueprint for understanding compound flood mechanisms and developing flood forecasting systems, especially for coastal regions. 

\section{Related Work}  
\textbf{Flood Datasets}. Monitoring floods presents a significant challenge due to unpredictable nature and potentially devastating consequences. Existing flood datasets can be broadly categorized into satellite image datasets~\citep{9460988, essd-14-1549-2022, 9882096, papagiannaki2022developing, xu2025floodcastbench, bonafilia2020sen1floods11} and time series monitoring datasets~\citep{xu2023impact,kabir2020deep, ADIKARI2021105136,RUMA2023100951,klingler2021lamah,essd-13-3847-2021,chagas2020camels}.
Satellite image datasets utilize remote sensing to capture the extent of surface water~\citep{rs16040656}. While effective in delineating flood-affected areas, this type of dataset~\citep{xu2025floodcastbench,bonafilia2020sen1floods11} often lacks temporal dynamics and information on the underlying hydrological and meteorological factors driving flood formation. 
Time series monitoring datasets typically utilize fixed monitoring stations to record hydrological-related data such as soil moisture and temperature. A prominent example within this category is the CAMELS-x family of datasets~\citep{addor2017camels,alvarez2018camels,coxon2020camels,chagas2020camels, essd-13-3847-2021}. For instance, CAMELS-BR~\citep{chagas2020camels} encompasses data from 3,679 gauges across Brazil. LamaH-CE~\citep{klingler2021lamah} provides daily and hourly time series data from 882 gauges, including runoff and catchment attributes. 
In~\citep{kirschstein2024merit}, LamaH-CE is used as a benchmark for flood forecasting, primarily focusing on water stream data across temporal and spatial aspects. In summary, these datasets primarily focus on general hydrological modeling. 
Analyzing compound floods, as highlighted in~\citep{nhess-20-2681-2020}, necessitates detailed data on rainfall, water levels, and groundwater, which are often limited in existing time series datasets.

\noindent \textbf{Data-Driven Methods for Forecasting}. 
High computational costs often constrain traditional physical models~\cite{paniconi2015physically,yin2023physic,brunner1997hec}. In contrast, data-driven AI excels at extracting complex, non-linear patterns directly from high-dimensional data, providing a scalable framework that can efficiently decode the spatial entanglement of multi-hazard drivers across diverse regions.
Deep learning methodologies have shown promise in this domain. Based on their core architectural designs, these methods applied to time series forecasting can be categorized as follows: \textit{MLP-based models}~\citep{chen2023tsmixer,zeng2023transformers,wang2024timemixer,pmlr-v235-lin24n,NEURIPS2024_bfe79983} leverage the capabilities of multilayer perceptrons for analyzing temporal sequences.
\textit{RNN-based methods}~\citep{lai2018modeling,salinas2020deepar,wang2018multilevel,qin2017dual,jhin2024addressing} are widely adopted in forecasting due to their inherent ability to model temporal dependencies within sequential data.
\textit{CNN-based methods}~\citep{wang2024timemixerpp,cheng2024convtimenet,luo2024moderntcn,wu2023timesnet,wang2023micn} employ convolutional operations to extract hierarchical features from time series data, enabling effective learning of underlying patterns and trends.
\textit{GNN-based methods}~\citep{wu2020connecting,wu2019graphwavenetdeepspatialtemporal,stemgnn,NEURIPS2022_7b102c90,FourierGNN,cai2023msgnet} utilize graph structures to model intricate relationships between different time series variables, enhancing forecasting accuracy.
\textit{Transformer-based methods}~\citep{wang2024timexer,liu2024itransformer, nie2023a, zhou2022fedformer,liu2022pyraformer, wu2021autoformer, zhou2021informer} have demonstrated remarkable performance in capturing long-range dependencies and complex temporal dynamics within time series data. \textit{LLM-based methods}~\citep{jin2024timellm,pan2024s,onefitsall,llmtime} explore the application of prompting and reprogramming techniques to align time series data with text embeddings for forecasting tasks.

\begin{table*}[t]
\centering
\caption{\textbf{Comparison with other datasets for flood forecasting.} N/A indicates Not Available. $^{*}$ marks datasets with Other Attributions. Climatic indices capture climate statistics (e.g., evapotranspiration). Land cover attributes describe surface materials. Soil attributes include properties such as porosity, while geological attributes refer to subsurface features. Anthropogenic influences cover human activities (e.g., discharges). Other catchment attributes include location, area, and topography. 
} 
\vspace{-0.3cm}
\label{tab:dataset_comparison}
\scalebox{0.95}{
    \begin{tabular}{llcccccl}
        \toprule
        Dataset & Time Span & Interval & Type & Gauges & Area(km$^2$)  & Public   & Other Attributions \\
     \midrule
     DarlingFlood
     & 1900-2018 & Daily & Flow & 12 & $3.5 \times 10^{4}$ &  No    & Rainfall \\
     \rowcolor{gray!20} SekongFlood
     & 1981-2013 & Daily &Flow & 8 & $2.8\times 10^{4}$ &  No & Rainfall \\
     BangladeshFlood
     & 1979-2013 & Daily & Stage & 24 & $1.5\times 10^{5}$ & No    & N/A \\
     \rowcolor{gray!20} Qi River
     & 1979-2020 & Hour & Flow & 7 & $7.1\times 10^{3}$  & No   & Rainfall  \\ 
     Tunxi basins
     & 1981-2007 & Hour & Flow  & 12 & N/A & No   & Rainfall \\ 
     \rowcolor{gray!20} CAMELS$^{*}$ 
     & 1989-2009 & Daily & Flow & 671 & $1.0\times 10^{4}$ & Yes   & Climatic Indices  \\
     CAMELS-CL$^{*}$
     & 1913-2018 & Daily & Flow & 516 & N/A  & Yes   & Land Cover Attributes\\
     \rowcolor{gray!20} CAMELS-GB$^{*}$
     & 1970-2015 & Daily & Flow  & 671 & $2.1\times 10^{5}$  & Yes  & Soil Attributes  \\ 
      CAMELS-BR$^{*}$
     & 1925-2024 & Daily  & Flow  &  4,025 & N/A  & Yes  & Geological Attributes  \\
     \rowcolor{gray!20} CAMELS-AUS$^{*}$
     & 1951-2014 & Daily & Flow &107 & $6.9\times 10^{5}$ & Yes   & Anthropogenic Influences \\
      LamaH-CE$^{*}$
     & 1951-2014 & Daily \& Hour & Flow & 859 & $1.7\times 10^{5}$ & Yes  & Other Catchment Attributes \\
    \midrule
    \rowcolor{gray!20}  {\dataset} & 1985-2024 & \begin{tabular}{@{}c@{}} Hour\end{tabular} & Stage & 2,452 & $6.7\times 10^{4}$  & Yes   & Rainfall, Groundwater, Human Control \\ 
    \bottomrule
    \end{tabular}
    }   
    \vspace{-0.2cm}
\end{table*}
\section{Methods} 
To enable systematic analysis of these interactions, we first construct an observational data framework, {\dataset}, that integrates heterogeneous data sources with explicit physical meaning into a coherent spatiotemporal representation.  Then, we perform spatiotemporal alignment to construct a coherent representation. After that, we apply data-driven AI analysis to reveal the underlying dependencies among various processes.

\subsection{Data Sources and Physical Interpretation}
{\dataset} is an observational time series dataset that characterizes the evolving state of a managed coastal hydrological system in South Florida. It aggregates measurements from 2,452 monitoring stations operated by the South Florida Water Management District (SFWMD)\footnote{https://www.sfwmd.gov/}, covering a spatial extent of 67,349~km$^2$ and a temporal range from 1985 to 2024. Because station availability varies over time, the dataset is organized into eight temporal splits that reflect the effective coverage period of individual monitoring locations.

The dataset integrates multiple physical and anthropogenic variables that jointly govern the emergence of compound flooding events~\citep{nhess-20-2681-2020}. These variables include surface water stage, rainfall, groundwater stage, coastal water levels, and records of human-controlled hydraulic operations. Together, they define the instantaneous state of the regional flood system by capturing external forcings, boundary conditions, internal storage, and active system controls.
Rainfall serves as the primary external forcing driving surface runoff and water accumulation.  Groundwater stage represents antecedent subsurface storage and reflects the land’s capacity to absorb incoming precipitation, which critically modulates flood severity during extreme rainfall events.  Coastal water levels, including tidal influences, act as dynamic boundary conditions for inland drainage. At several monitoring stations, these coastal effects are inherently embedded in the observed water stage due to direct hydraulic connectivity with the ocean. In addition to natural drivers, {\dataset} explicitly incorporates records of human-controlled hydraulic operations, including pumps and gates. Pumps enable water transfer from lower to higher elevations, while gates regulate downstream flow from higher to lower elevations. In South Florida, where an extensive network of canals, lakes, storage areas, and control structures actively manages water distribution, such operational interventions directly alter system dynamics and flood propagation pathways. Representing these controls is therefore essential for capturing the coupled human–natural processes underlying compound flood behavior. 

A comparison between {\dataset} and existing hydrological datasets is provided in Table~\ref{tab:dataset_comparison}. Compared to CAMELS-x~\citep{addor2017camels,alvarez2018camels,coxon2020camels,chagas2020camels,essd-13-3847-2021}, which emphasizes large-sample hydrology across diverse regions, {\dataset} focuses on a flood-prone coastal area where compound flooding is a dominant hazard. In contrast to BangladeshFlood~\citep{RUMA2023100951}, {\dataset} incorporates a broader set of interacting drivers, capturing both temporal evolution and spatial heterogeneity relevant to compound flood processes.

\subsection{Spatiotemporal Alignment}
To facilitate the construction of AI-ready datasets, we provide detailed information on aligning heterogeneous observational records.

\noindent \textbf{The Observational Network}. 
The raw observations underlying {\dataset} are sourced from DBHYDRO\footnote{\url{https://apps.sfwmd.gov/dbhydroInsights/\%23/homepage}}, an environmental data repository maintained by SFWMD. 
The raw data stream, originally encompassing 3,731 sensors, underwent a rigorous spatiotemporal density filtration protocol. We prioritized stations with long-term data continuity and high relevance to cross-boundary hydraulic interactions, resulting in a finalized network of 2,452 high-fidelity nodes. This multi-modal array comprises water stage ($n=993$), precipitation ($n=349$), and groundwater levels ($n=582$), integrated with the active control signals of 99 pump stations and 429 hydraulic gates. By synthesizing these disparate sensing modalities, the repository provides a comprehensive digital pulse of the region's hydrological state.
Each station provides localized observations of distinct system components. Together, these measurements form a distributed sensing network that captures external forcings, internal storage states, boundary conditions, and human-controlled interventions within the regional flood system. In our framework, water stage, rather than streamflow, is used here because streamflow approaches zero and is hard to measure in many catchments, such as lakes and wetlands. A summary of data types and their physical interpretations is provided in Table~\ref{tab:data_type}. 

\noindent \textbf{Temporal Alignment}. 
Raw observations in the DBHYDRO are recorded at native breakpoint frequencies, which vary substantially across stations and variables, ranging from sub-hourly to irregular event-based recordings. While this high-resolution logging preserves fine-scale dynamics, it results in heterogeneous and asynchronous time series that are not directly comparable across space or variables. Moreover, observational gaps and uneven station lifespans introduce additional temporal inconsistency.
To construct a unified system-level representation, we align all observations to an hourly time grid. For each station and variable, measurements within each hourly interval are aggregated using their mean value, yielding a temporally consistent snapshot. 
This hourly resolution balances the need to capture flood-relevant dynamics with the requirement for long-term temporal coverage across stations. 
Due to heterogeneous data availability, the aligned time series are organized into eight contiguous temporal splits\footnote{Detailed split information is reported in Appendix Table~\ref{tab:splits}.}, each corresponding to a contiguous period with sufficient station coverage. 

\noindent \textbf{Handling Missing Observations}. 
Temporal alignment inevitably introduces missing values where observations are unavailable. 
To ensure the integrity of the physical signals during this transition, we applied differentiated interpolation strategies based on the variables' nature: continuous field and event-based sparse representation. 
Surface water stage and groundwater level are treated as continuous state variables that evolve smoothly over time, and missing values are filled using linear interpolation.  In contrast, rainfall and hydraulic control records (pumps and gates) represent stochastic external events; missing entries are therefore interpreted as the absence of events and filled with zero.

\subsection{Data-Driven AI Analysis}
We investigate the extent to which data-driven models can capture the coupled dynamics of a human-regulated hydrological system using large-scale observational data. 
By reframing water stage forecasting from a purely predictive task into a system identification problem, we aim to reveal the underlying spatiotemporal dependencies linking surface water dynamics, subsurface processes, meteorological forcing, and human interventions.

\noindent \textbf{Scientific Objective}.  We examine whether AI models can learn hydrological dynamics and forecast water stages from multi-factor observational data. Specifically, we aim to understand how surface water levels evolve in response to meteorological forcing, subsurface groundwater interactions, and human-operated hydraulic controls, such as pumps and gates.

We formulate the hydrological system as a spatio-temporal coupled network driven by both natural processes and human control signals.  Given water stage time series data $\mX \in \mathbb{R}^{N\times L}$ from $N$ water monitoring stations, the forecasting objective is to predict the water stage values for the next $T$ time steps, denoted as $\mY \in \mathbb{R}^{N\times T}$, using a fixed look-back window of length $L$. In addition to the surface water stage, the system also has access to additional time series information, including groundwater levels $\mX_w \in \mathbb{R}^{N_w\times L}$ (from $N_w$ stations), rainfall $\mX_r \in \mathbb{R}^{N_r\times L}$ (from $N_r$ stations), pump control data $\mX_p \in \mathbb{R}^{N_p\times L}$ (from $N_p$  stations), gate control data $\mX_g \in \mathbb{R}^{N_g\times L}$ (from $N_g$ stations), and location information for the monitoring stations.  From the data-driven model perspective, this formulation treats water stage evolution as the outcome of interacting latent processes operating across space and time.  For the primary benchmark experiments aimed at fair comparison across different modeling approaches, we focus on water stage forecasting as the supervised prediction target.  The additional modalities are regarded as external drivers that can potentially enrich system understanding and improve predictive performance, and their integration is explored in extended experimental settings.

\begin{table}[t]
  \caption{The summary of the data type in {\dataset}.}
  \vspace{-0.3cm}
  \label{tab:data_type}
  \scalebox{0.95}{
    \begin{tabular}{lccl}
        \toprule
     Type        &  Stations    &  Unit   & Description  \\
     \midrule
     Water  &   993  & Feet  & Water Stage  \\ 
     \rowcolor{gray!20} Groundwater &   582  & Feet  &  Stage of Groundwater \\ 
     Rainfall &  349 &  Inches &  Rainfall \\ 
     \rowcolor{gray!20} Pump &  99 &   RPM & Rotational Speed \\ 
     Gate &   429 &   Feet & Opening Level \\
    \bottomrule
    \end{tabular}
    }   
    \vspace{-0.5cm}
\end{table}

\noindent \textbf{Data-Driven Models}. To cover a broad range of inductive biases, we benchmark six categories of time series forecasting architectures, including MLP, CNN, RNN, GNN, Transformer, and LLM-based models.  We select two advanced methods as representative examples for each of these categories. Additionally, for the first four classical architectures, we also implement a basic foundational architecture. The specific advanced methods we evaluate are: 1. MLP-Based Models: NLinear~\citep{zeng2023transformers}, TSMixer~\citep{chen2023tsmixer}, 2. CNN-Based Models: ModernTCN~\citep{luo2024moderntcn}, TimesNet~\citep{wu2023timesnet}, 3. RNN-Based Models: DeepAR~\citep{salinas2020deepar}, DilatedRNN~\citep{chang2017dilated}, 4. GNN-Based Models: FourierGNN~\citep{FourierGNN}, StemGNN~\citep{stemgnn}, 5. Transformer-Based Models: PatchTST~\citep{nie2023a}, iTransformer~\citep{liu2024itransformer}, 6. LLM-Based Models: GPT4TS~\citep{onefitsall}, AutoTimes~\citep{liu2024autotimes}. We follow the source code of NeuralForecast~\footnote{https://github.com/Nixtla/neuralforecast} for the implementation of these approaches\footnote{The summary of these methods can be found in Appendix~\ref{appendix:method}.}. 

\subsection{Experimental Design}
We design a series of controlled experiments to examine whether existing machine learning models can capture the spatiotemporal and multi-factor characteristics of compound flooding systems. Rather than optimizing for a single benchmark score, our experimental design aims to answer three scientific questions: 
\textit{(i) whether current AI models can effectively learn spatiotemporal dependencies under multi-factor factors}, 
\textit{(ii) which input factors are most informative for compound flooding prediction}, and 
\textit{(iii) the relative importance of temporal versus spatial context in forecasting performance}.

\noindent \textbf{Spatial Scope and Computational Constraints}. 
Due to the substantial memory and training costs associated with large-scale models, particularly LLM-based approaches, we adopt a two-level evaluation strategy. First, we conduct comprehensive benchmarking within three hydrologically relevant regions selected based on observed flood occurrences. The three areas are selected based on the flood observation data presented in Figure~\ref{fig:loc}, which visualizes these flood locations alongside the selected areas. These regions represent areas where compound flooding effects are most pronounced. Second, for the full dataset, we report results for four foundational methods to assess large-scale generalization while maintaining computational feasibility.

\noindent \textbf{Factor and Context Ablations}. 
To isolate the contribution of different factors, we conduct targeted ablation studies. These include removing or selectively incorporating groundwater, rainfall, and human control signals, as well as varying the spatial extent of input stations and the temporal length of historical context. By systematically controlling these factors, we assess how different models leverage temporal accumulation, spatial coupling, and cross-modal information in learning compound flooding dynamics.

\noindent \textbf{Temporal Splits and Forecasting Horizons}. 
To evaluate generalization across hydrological regimes, we apply temporally disjoint data splits, using the last year for testing, the second-to-last year for validation, and all preceding data for training. A fixed look-back window of two days is used for all models, while prediction horizons of one, three, five, and seven days are evaluated to assess short- and medium-term forecasting capability.

\subsection{Evaluation Criteria}
The evaluation is designed to reflect both overall predictive fidelity and sensitivity to hydrologically critical extreme events. To quantify general predictive accuracy, we follow standard time series forecasting practices to report the Mean Absolute Error (MAE) and Mean Squared Error (MSE), which measure deviations between predicted and observed water stage values across all time steps.  

In addition, to explicitly evaluate performance on extreme flood-related events, we also employ the Symmetric Extremal Dependence Index (SEDI)~\citep{han2024cra5,xu2024extremecast}, following prior work in flood and climate forecasting~\citep{han2024far}. The formulation of SEDI is as follows:
\begin{equation}
    \text{SEDI}(p) = \frac{|\hat{\mY}<V_{1-\frac{p}{2}} \& {\mY}<V_{1-\frac{p}{2}} |+|\hat{\mY}>V_{\frac{p}{2}} \& {\mY}>V_{\frac{p}{2}}|}{|{\mY}<V_{1-\frac{p}{2}}|+|{\mY}>V_{\frac{p}{2}}|},
\end{equation}
where $|\cdot|$ means the number of the true values satisfying the condition, $\hat{\mY}$ is the forecasting results, $p$ is the quantile of the threshold, and $V_{1-\frac{p}{2}}, V_{\frac{p}{2}}$ are the lower and upper threshold of top and worst $\frac{p}{2}$ percent, respectively.  SEDI focuses on the co-occurrence of extreme predictions and observations, thereby emphasizing the model’s responsiveness to tail events. Specifically, extreme conditions are defined using quantile-based thresholds (e.g., the 95th and 5th percentiles of the observed values), and each time step is classified as either normal or extreme.  A higher SEDI score indicates stronger dependence between predicted and observed extremes, reflecting improved capability in capturing flood-relevant dynamics. 

\subsection{Experimental Setup}
In this paper, the default look-back window size is two days, and the forecast window length is a list of $[1,3,5,7]$ days. For the SEDI metric, we report the average performance across four forecast settings. Except for Table~\ref{tab:avg_part_all_msemae_new}, MAE and MSE results are reported based on the average of four forecast window lengths.

\begin{figure*}[t]
    \centering
    \scalebox{0.85}{
    \begin{subfigure}[h]{0.2\textwidth}
        \includegraphics[width=\textwidth]{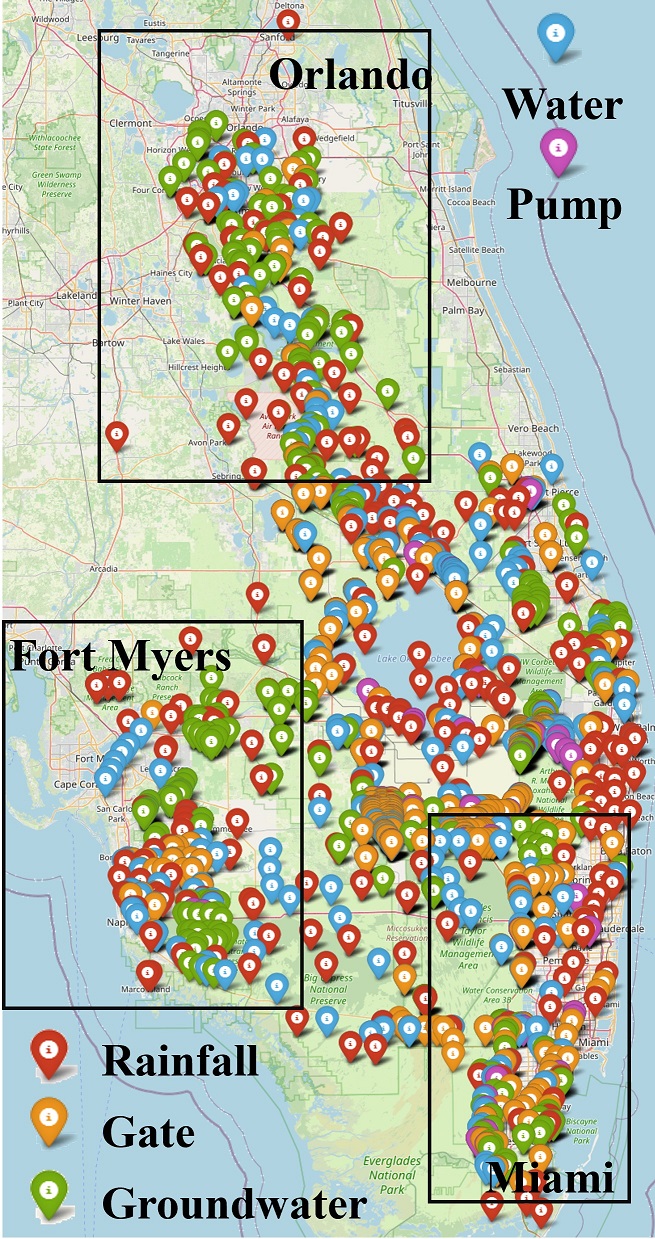}
        \caption{}
        \label{fig:loc}
    \end{subfigure}
    \hspace{0.7cm}
    \begin{subfigure}[h]{0.2\textwidth}
        \includegraphics[width=\textwidth]{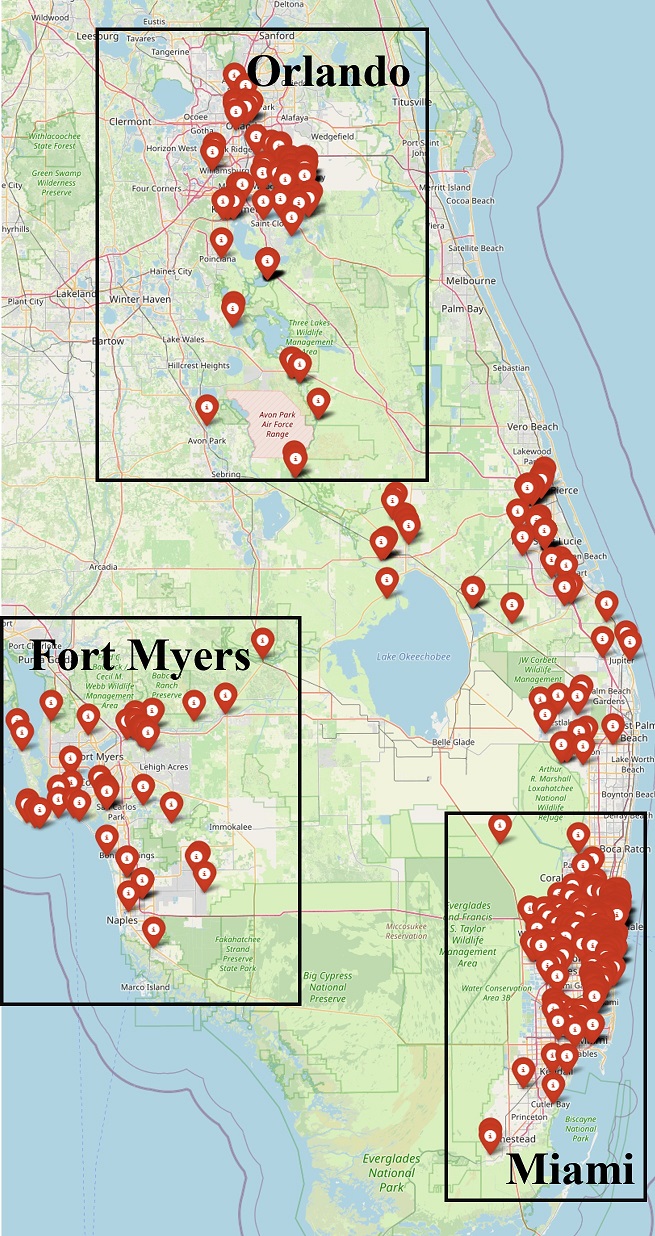}
        \caption{}
        \label{fig:loc_flood}
    \end{subfigure}
    
    \hspace{0.7cm}
    
     \begin{subfigure}[h]{0.20\textwidth}
        \includegraphics[width=\textwidth, trim=20 0 60 0, clip]{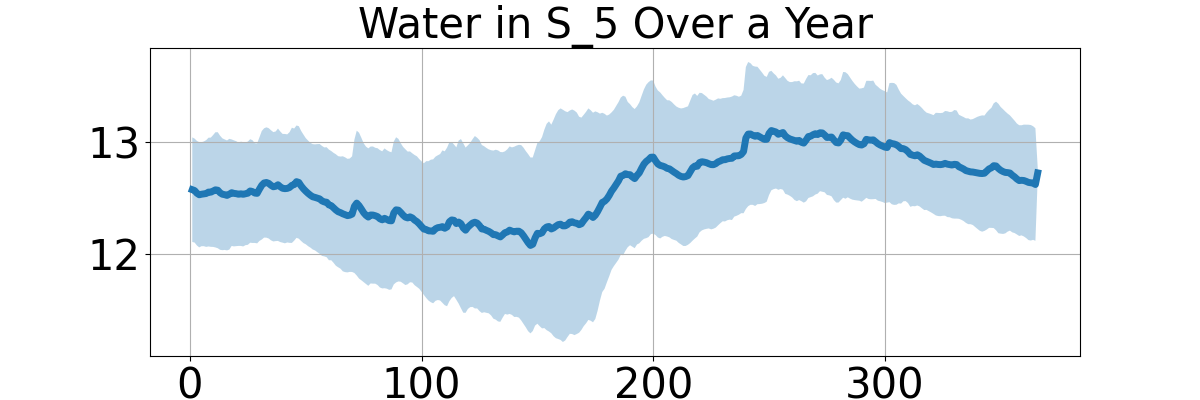}
        \includegraphics[width=\textwidth, trim=20 0 60 0, clip]{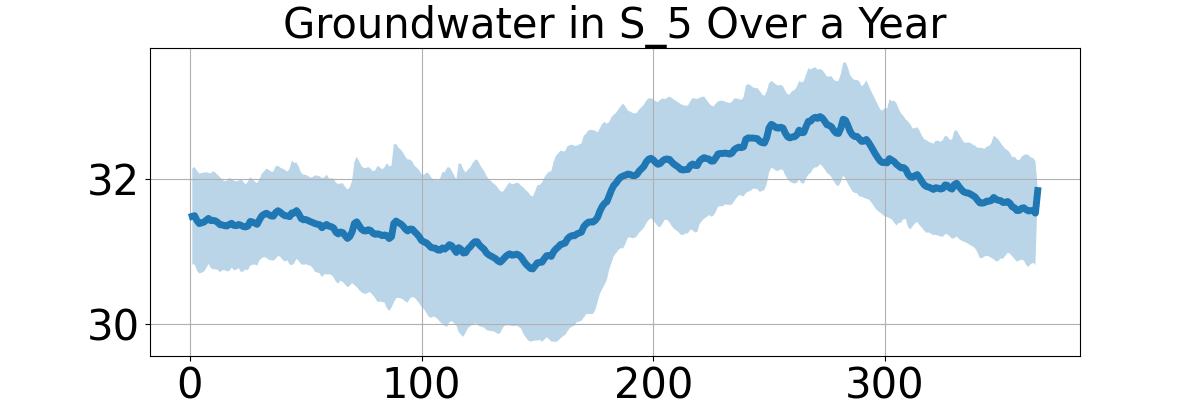}
        \includegraphics[width=\textwidth, trim=20 0 60 0, clip]{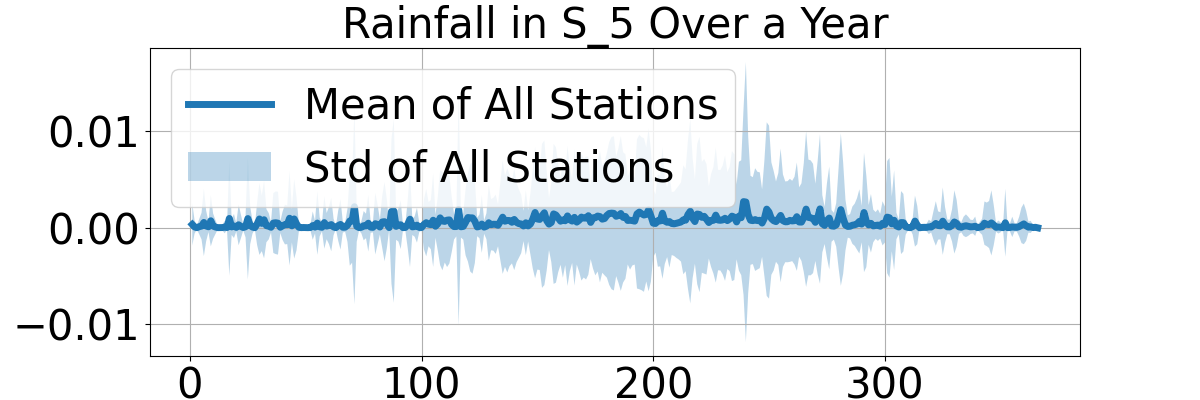}
        \includegraphics[width=\textwidth, trim=20 0 60 0, clip]{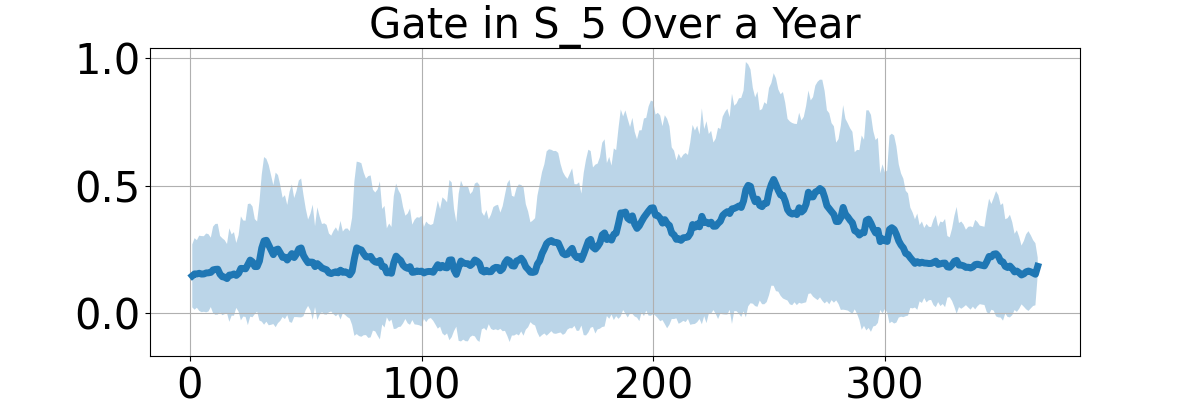}
        \includegraphics[width=\textwidth, trim=20 0 60 0, clip]{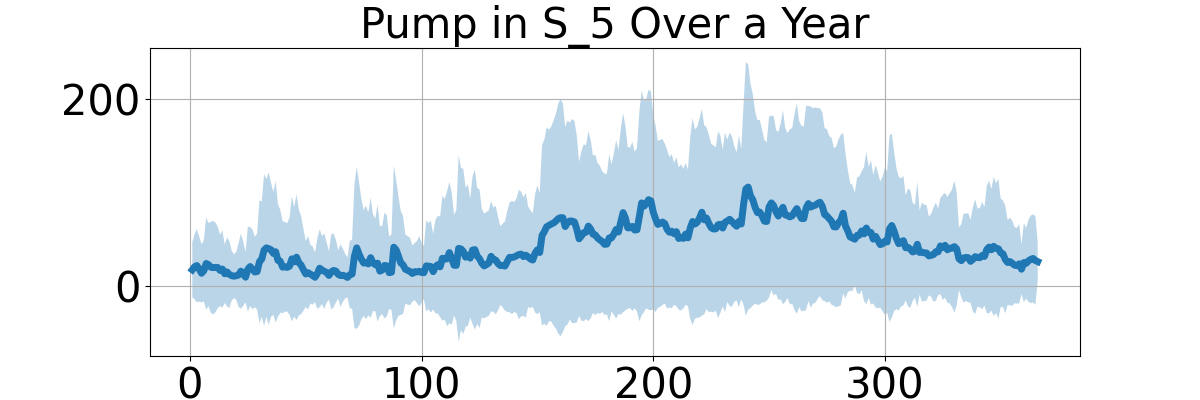}
        \caption{}
        \label{fig:key_pattern_S6}
    \end{subfigure}
    \hspace{0.7cm}
    
    \begin{subfigure}[h]{0.20\textwidth}
        \includegraphics[width=\textwidth, trim=20 0 60 0, clip]{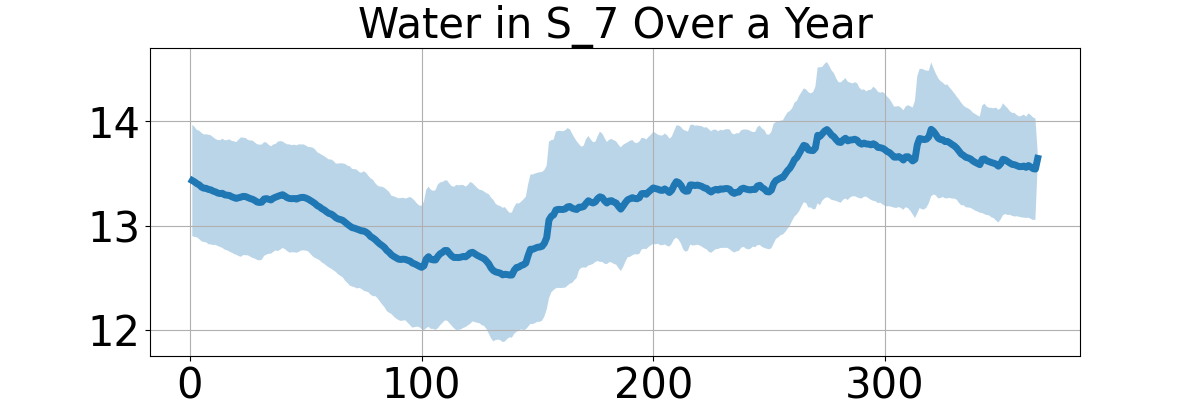}
        \includegraphics[width=\textwidth, trim=20 0 60 0, clip]{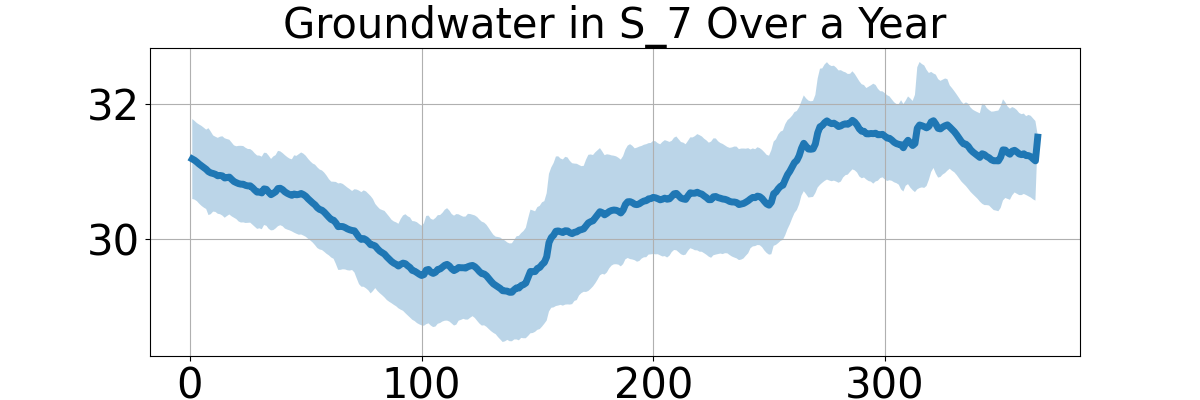}
        \includegraphics[width=\textwidth, trim=20 0 60 0, clip]{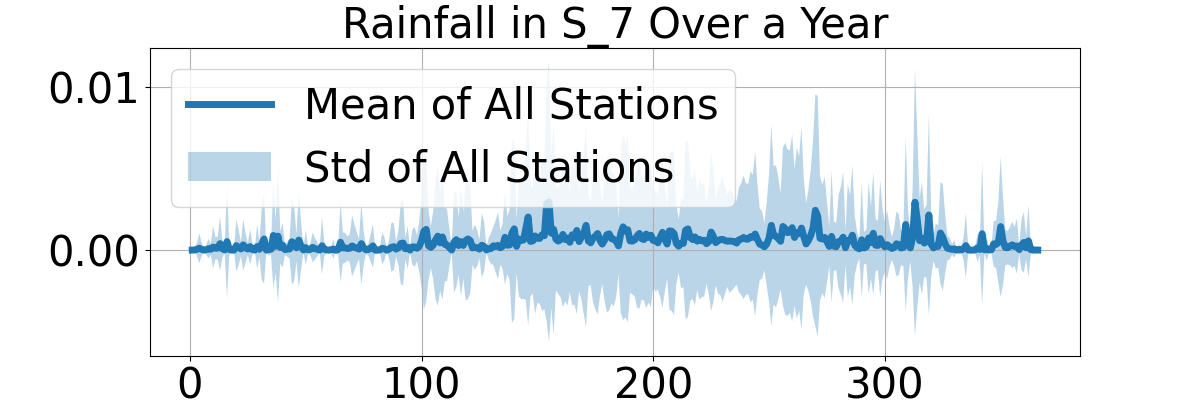}
        \includegraphics[width=\textwidth, trim=20 0 60 0, clip]{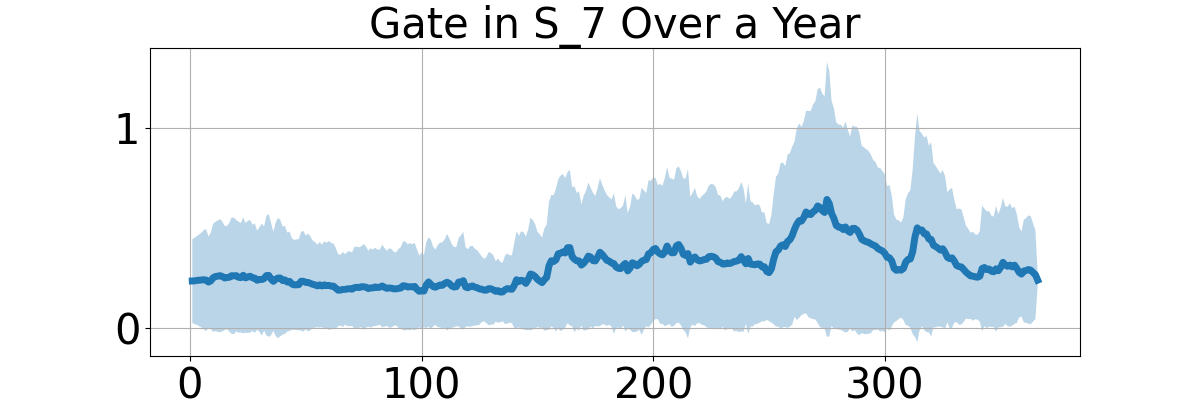}
        \includegraphics[width=\textwidth, trim=20 0 60 0, clip]{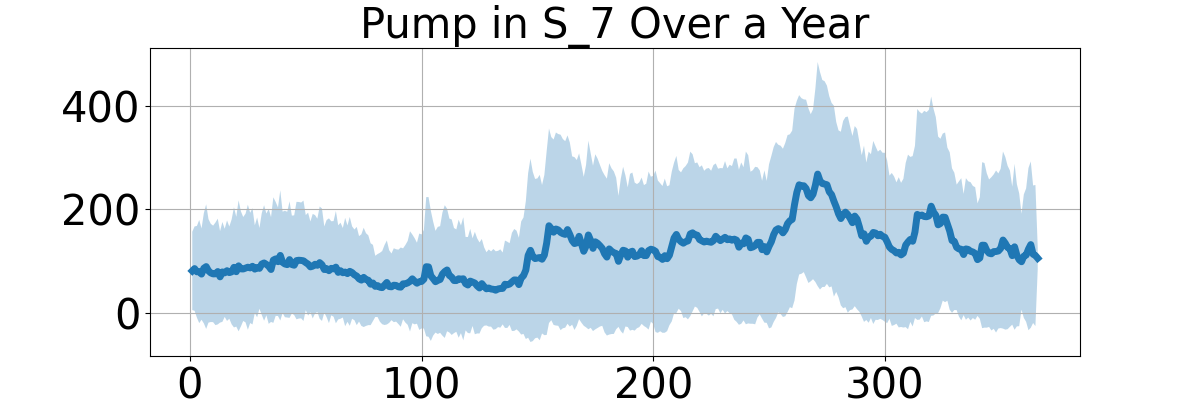}
        \caption{}
        \label{fig:key_pattern_S7}
    \end{subfigure}
    }
    \vspace{-0.4cm}
    \caption{(a). Location distribution of monitor stations. (b). Flood observation location distribution. In (a) and (b), we highlight three interest parts: Orlando, Fort Myers, and Miami. (c) and (d). Temporal patterns of key features over a year. The x-axis is the number of days in one year. }
    \label{fig:spatial_temporal}
    \vspace{-0.4cm}
\end{figure*}

\section{Results}
\label{sec:res}
\subsection{Dataset Analysis \&  Visualization}
To establish system-level intuition and verify the physical consistency of {\dataset}, we conduct a comprehensive series of visualization-based analyses from both spatial and temporal perspectives. These analyses serve as sanity checks to examine whether the observed patterns are coherent across space and time and consistent with established hydrological processes. 

\noindent \textbf{Spatial Distribution.} Figure~\ref{fig:loc} presents the geographic distribution of all monitoring stations, which are predominantly deployed along the complex river network of South Florida and radiate outward from Lake Okeechobee. The staggered spatial layout of hydrological, groundwater, and rainfall stations enables joint spatial reasoning across different components of the hydrological system. For instance, water level variations at a given site are expected to be influenced by nearby precipitation patterns as well as local groundwater conditions, which reflect subsurface water storage capacity.  
To further contextualize the dataset, we overlay observed flood events recorded in 2008 and from 2020 to 2023 in Figure~\ref{fig:loc_flood}, based on reports from the South Florida Water Management District (SFWMD)~\citep{OfficeResilienceSFWMD}. Notably, these flood events are primarily concentrated in urbanized regions, highlighting the interaction between natural hydrological processes and human-modified landscapes.

\noindent \textbf{Temporal Pattern}. Figure~\ref{fig:spatial_temporal} illustrates the average annual temporal patterns across all monitoring stations, using data from split S5 and S7 as a representative example. From these patterns, we observe a strong correlation between groundwater level and water level data. The water stage generally rises from approximately the 150th to the 300th day of the year, followed by a gradual decrease. According to the National Weather Service (NWS), Florida's climate is characterized by distinct dry and rainy seasons, with the latter typically spanning from May to October. This aligns well with the observed data patterns in our dataset. Taking rainfall as a reference, we note that increases in water level tend to correspond with rainfall events, such as the rainfall peak after the 300th day and the subsequent rise in water level. In addition, human control activities on hydraulic structures appear to influence water level changes in response to rainfall. We can infer that human intervention has played a role in mitigating potential flooding.

\subsection{Observations \& Discovery}
\label{sec:discovery}
We organize our observations around the core scientific questions concerning the modeling of compound flooding systems.

\noindent \textbf{Q1: Can existing machine learning models effectively capture multi-factor spatiotemporal dynamics?}  

Across all benchmark experiments, we observe that current machine learning models exhibit uneven capability in capturing the multi-factor and spatiotemporal structure of the hydrological system. 
Table~\ref{tab:avg_part_all_msemae_new} reports the average MSE and MAE results across three regions and eight data splits. The top-performing methods include PatchTST, iTransformer, MLP, NLinear, and AutoTimes. PatchTST and iTransformer achieve the best MAE, while MLP, NLinear, and AutoTimes perform best under MSE. In general, a small MAE but large MSE suggests high accuracy for most points with occasional large errors, whereas the opposite indicates greater stability but lower pointwise accuracy. Table~\ref{tab:avg_part_all_msemae_new} also shows model performance on extreme cases; in contrast, ModernTCN and NLinear demonstrate competitive performance on extreme-event metrics. We also observe similar results for performance across all stations in Appendix~\ref{appendix:extend_res}.

From our observations, the MLP achieves near-best performance across almost all experiments, despite its relatively concise model structure. This suggests that existing methods that emphasize temporal dependency may fail to adequately capture the interactions among multiple driving factors. Furthermore, these results reveal a trade-off in model biases that certain architectures favor overall system averages, whereas others are specifically tuned to the patterns of correlated extremes.

\begin{table*}[t]
\centering
\captionof{table}{Benchmark of average MAE, MSE, and SEDI results on three interest areas across 8 splits. The best and second results are shown in bold font and underlined, respectively. The best performances are in \textcolor{cyan}{\textbf{bold}}, and the second-best method is \textcolor{orange}{\underline{underlined}}.}
\vspace{-0.3cm}
\label{tab:avg_part_all_msemae_new}
\scalebox{0.90}{
    \begin{tabular}{l|ccccc|ccccc|ccc}
        \toprule
        & \multicolumn{5}{|c}{\textbf{MAE$\downarrow$}}
       & \multicolumn{5}{|c}{\textbf{MSE}$\downarrow$}
       & \multicolumn{3}{|c}{\textbf{SEDI}$\uparrow$} \\
       \cmidrule(r){2-6} \cmidrule(r){7-11}  \cmidrule(r){12-14}
         Methods & 1D & 3D  & 5D   & 7D & Avg. & 1D & 3D  & 5D   & 7D & Avg.  & 10\%  & 5\%  & 1\%  \\
    \midrule
    \rowcolor{gray!20} \multicolumn{14}{c}{\textbf{MLP-Based Models}} \\
      MLP            & 0.0788 & 0.1351 & 0.1764 & 0.2063 & 0.1492 & \textcolor{orange}{\underline{0.0521}} & \textcolor{cyan}{\textbf{0.1029}} & \textcolor{cyan}{\textbf{0.1432}} & \textcolor{cyan}{\textbf{0.1707}} & \textcolor{cyan}{\textbf{0.1172}} & 0.5302 & 0.3952 & 0.1721 \\
      TSMixer        & 0.0928 & 0.1442 & 0.1816 & 0.2126 & 0.1578 & 0.0648 & 0.1132 & 0.1566 & 0.1903 & 0.1283 & 0.5927 & 0.4639 & 0.2274 \\
      NLinear        & 0.0817 & 0.1373 & 0.1769 & 0.2082 & 0.1510 & 0.0556 & \textcolor{orange}{\underline{0.1105}} & 0.1547 & 0.1841 & \textcolor{orange}{\underline{0.1262}} & 0.6097 & 0.4825 & 0.2513 \\
    \midrule
    \rowcolor{gray!20} \multicolumn{14}{c}{\textbf{CNN-Based Models}} \\
      TCN            & 0.1792 & 0.2281 & 0.2697 & 0.3088 & 0.2465 & 0.3722 & 0.4647 & 0.4173 & 0.5386 & 0.4482 & 0.2020 & 0.1449 & 0.0983 \\
      ModernTCN      & 0.0798 & 0.1441 & 0.1891 & 0.2248 & 0.1594 & 0.0681 & 0.2443 & 0.2808 & 0.2723 & 0.2164 & \textcolor{orange}{\underline{0.6279}} & \textcolor{cyan}{\textbf{0.5130}} & \textcolor{cyan}{\textbf{0.2828}} \\
      TimesNet       & 0.0983 & 0.1528 & 0.1927 & 0.2267 & 0.1676 & 0.0704 & 0.1358 & 0.1829 & 0.2241 & 0.1533 & 0.6079 & 0.4880 & 0.2509 \\
    \midrule
    \rowcolor{gray!20} \multicolumn{14}{c}{\textbf{RNN-Based Models}} \\
      LSTM           & 0.1182 & 0.1821 & 0.2232 & 0.2576 & 0.1953 & 0.1339 & 0.1985 & 0.2348 & 0.2873 & 0.2136 & 0.4427 & 0.3062 & 0.0867 \\
      DeepAR         & 0.1178 & 0.1837 & 0.2247 & 0.2596 & 0.1964 & 0.1230 & 0.1954 & 0.2389 & 0.2691 & 0.2066 & 0.4271 & 0.2852 & 0.0663 \\
      DilatedRNN     & 0.0919 & 0.1573 & 0.2022 & 0.2374 & 0.1722 & 0.1033 & 0.1672 & 0.2341 & 0.2591 & 0.1909 & 0.5536 & 0.4179 & 0.1730 \\
    \midrule
    \rowcolor{gray!20} \multicolumn{14}{c}{\textbf{GNN-Based Models}} \\
      GCN            & 0.1696 & 0.2006 & 0.2504 & 0.2799 & 0.2251 & 7.2299 & 1.5645 & 2.1341 & 1.0374 & 2.9915 & 0.3662 & 0.2582 & 0.1210 \\
      StemGNN        & 0.1332 & 0.2181 & 0.3153 & 0.3570 & 0.2559 & 0.1632 & 0.2543 & 0.4189 & 0.4716 & 0.3270 & 0.4350 & 0.2993 & 0.0985 \\
      FourierGNN     & 0.0921 & 0.1503 & 0.1930 & 0.2280 & 0.1658 & 0.0768 & 0.1416 & 0.2071 & 0.2125 & 0.1595 & 0.4657 & 0.3647 & 0.1645 \\
    \midrule
    \rowcolor{gray!20} \multicolumn{14}{c}{\textbf{Transformer-Based Models}} \\
      PatchTST       & \textcolor{cyan}{\textbf{0.0741}} & \textcolor{orange}{\underline{0.1316}} & \textcolor{cyan}{\textbf{0.1719}} & \textcolor{orange}{\underline{0.2044}} & \textcolor{cyan}{\textbf{0.1455}} & 0.0531 & 0.1132 & 0.1566 & 0.1903 & 0.1283 & 0.6254 & 0.5043 & 0.2711 \\
      iTransformer   & \textcolor{orange}{\underline{0.0756}} & \textcolor{cyan}{\textbf{0.1314}} & \textcolor{orange}{\underline{0.1722}} & \textcolor{cyan}{\textbf{0.2041}} & \textcolor{orange}{\underline{0.1458}} & \textcolor{cyan}{\textbf{0.0253}} & 0.1112 & 0.1583 & 0.1911 & 0.1284 & \textcolor{cyan}{\textbf{0.6292}} & \textcolor{orange}{\underline{0.5077}} & \textcolor{orange}{\underline{0.2705}} \\
    \midrule
    \rowcolor{gray!20} \multicolumn{14}{c}{\textbf{LLM-Based Models}} \\
      GPT4TS         & 0.1256 & 0.1521 & 0.1911 & 0.2247 & 0.1734 & 0.0966 & 0.1410 & 0.1847 & 0.2245 & 0.1617 & 0.6163 & 0.4924 & 0.2566 \\
      AutoTimes      & 0.0846 & 0.1362 & 0.1752 & 0.2062 & 0.1505 & 0.0584 & 0.1125 & \textcolor{orange}{\underline{0.1522}} & \textcolor{orange}{\underline{0.1823}} & 0.1263 & 0.6054 & 0.4794 & 0.2476 \\
     \bottomrule
    \end{tabular}
    }
    \vspace{-0.1cm}
\end{table*}

\begin{table}[t]
\centering
\captionof{table}{Input factors ablation study on S6. All, G, R, and C represent all factors: groundwater, rainfall, and human control(pump and gate). The best performances are in \textcolor{cyan}{\textbf{bold}}, and the second-best method is \textcolor{orange}{\underline{underlined}}.} 
    \vspace{-0.3cm}
\label{tab:avg_inputs_msemae}
\scalebox{0.93}{
    \begin{tabular}{llccc@{\hspace{0.25cm}}c}
        \toprule
       &     Models      & w/ All    & w/o G  & w/o R   & w/o C \\
    \midrule
    \multirow{5}{*}{\rotatebox[origin=c]{90}{MAE}} &  iTransformer   & 0.1406	&	0.1407	&	0.1406	&	\textcolor{orange}{\underline{0.1405}}	\\
                        &   \cellcolor{gray!20} PatchTST   & \cellcolor{gray!20} \textcolor{cyan}{\textbf{0.1376}}	&	\cellcolor{gray!20} 0.1391	&	\cellcolor{gray!20} 0.1378	&	\cellcolor{gray!20} \textcolor{orange}{\underline{0.1377}}	\\
                        &   TSMixer   & 0.1596	&	\textcolor{orange}{\underline{0.1419}}	&	0.2523	&	0.2111	\\
                        &  \cellcolor{gray!20} NLinear   & \cellcolor{gray!20} 0.1546	&	\cellcolor{gray!20} 0.1577	&	\cellcolor{gray!20} 0.1483	&	\cellcolor{gray!20} 0.1540	\\
                        &   TimesNet & 0.1642	&	0.1599	&	0.1662	&	0.1650	 \\
    \midrule
    \multirow{5}{*}{\rotatebox[origin=c]{90}{MSE}} &  \cellcolor{gray!20} iTransformer   & \cellcolor{gray!20} \textcolor{orange}{\underline{0.0953}}	&	\cellcolor{gray!20} 0.0960	&	\cellcolor{gray!20} \textcolor{cyan}{\textbf{0.0949}}	&	\cellcolor{gray!20} 0.0954	\\
                        &   PatchTST   & \textcolor{orange}{\underline{0.0917}}	&	0.0932	&	0.0927	& \textcolor{cyan}{\textbf{0.0916}} \\
                        &   \cellcolor{gray!20} TSMixer   & \cellcolor{gray!20} 0.1080	&	\cellcolor{gray!20} 0.0946	&	\cellcolor{gray!20} 0.4061	&	\cellcolor{gray!20} 0.2698	\\
                        &   NLinear   & 0.0992	&	0.1011	&	0.0966	&	0.0984	\\
                        &   \cellcolor{gray!20} TimesNet &\cellcolor{gray!20}  0.1188	&\cellcolor{gray!20} 	0.1154	&\cellcolor{gray!20} 	0.1226	&\cellcolor{gray!20} 	0.1202	 \\
    \hline
    \hline
     &     Models        & w/o GR  & w/o RC &  w/o GC & w/o GRC \\
    \midrule
    \multirow{5}{*}{\rotatebox[origin=c]{90}{MAE}} &  iTransformer   &	0.1411	&	0.1410	&	\textcolor{cyan}{\textbf{0.1402}}	&	0.1411 \\
                        &   \cellcolor{gray!20} PatchTST   &	\cellcolor{gray!20} 0.1398	&	\cellcolor{gray!20} 0.1396	&	\cellcolor{gray!20} 0.1385	&\cellcolor{gray!20} 	0.1421		  \\
                        &   TSMixer   &	\textcolor{cyan}{\textbf{0.1418}}	&	0.2807	&	0.1423	&	0.1422\\
                        &   \cellcolor{gray!20} NLinear   &	\cellcolor{gray!20} 0.1485	&\cellcolor{gray!20} 	\textcolor{orange}{\underline{0.1450}}	&	\cellcolor{gray!20} 0.1579	&	\cellcolor{gray!20}\textcolor{cyan}{\textbf{0.1435}}\\
                        &   TimesNet &	0.1592	&	0.1656	&	\textcolor{cyan}{\textbf{0.1578}}	&	\textcolor{orange}{\underline{0.1580}}\\
    \midrule
    \multirow{5}{*}{\rotatebox[origin=c]{90}{MSE}} &  \cellcolor{gray!20} iTransformer   &	\cellcolor{gray!20} 0.0957	&	\cellcolor{gray!20} 0.0956	&	\cellcolor{gray!20} \textcolor{cyan}{\textbf{0.0949}}	&\cellcolor{gray!20}	0.0962\\
                        &   PatchTST   &	0.0950	&	0.0938	&	0.0923	&	0.0971	  \\
                        &  \cellcolor{gray!20} TSMixer   &	\cellcolor{gray!20}0.0946	&\cellcolor{gray!20}	0.6697	&	\cellcolor{gray!20}\textcolor{orange}{\underline{0.0943}}	&	\cellcolor{gray!20}\textcolor{cyan}{\textbf{0.0941}}		  \\
                        &   NLinear   &	0.0973	&	\textcolor{orange}{\underline{0.0954}}	&	0.1006	&	\textcolor{cyan}{\textbf{0.0947}}		  \\
                        &  \cellcolor{gray!20} TimesNet &	\cellcolor{gray!20}0.1145	&\cellcolor{gray!20}	0.1237	&	\cellcolor{gray!20}\textcolor{cyan}{\textbf{0.1111}}	&	\cellcolor{gray!20}\textcolor{orange}{\underline{0.1123}} \\
    
     \bottomrule
    \end{tabular}
    }   
    \vspace{-0.3cm}
\end{table}

\noindent \textbf{Q2: Do average accuracy metrics reflect performance on extreme events (e.g., floods)?}  

We find that improvements in MAE and MSE do not fully correspond to improved extreme-event prediction. Models such as ModernTCN achieve strong SEDI performance despite weaker average accuracy, indicating a decoupling between mean behavior and tail dynamics. This highlights the necessity of evaluating compound flooding models using metrics that explicitly target extreme events, as average error measures alone may obscure critical system-level behaviors relevant to flood risk assessment.

\noindent \textbf{Q3: Which factors are most informative for compound flooding prediction?}  

To quantify the contribution of diverse drivers to compound flooding predictability, we conducted a systematic ablation study across five representative architectures. These models are categorized into two paradigms: channel-independent (NLinear, PatchTST) and channel-dependent (iTransformer, TimesNet, TSMixer). While the former treats all variables as a unified input pool, the latter variants designate water stage as the primary supervised target, treating auxiliary factors as exogenous covariates.

By systematically isolating factors including rainfall ($R$), anthropogenic control signals ($C$), and groundwater ($G$), we reveal a non-uniform contribution of auxiliary inputs that is deeply tied to model inductive biases. 
As shown in Table~\ref{tab:avg_inputs_msemae}, a distinct bifurcation emerges that, while NLinear and TSMixer reach a performance plateau using water stage data alone, architectures like iTransformer and PatchTST demonstrate a superior capacity for multimodal synergy, showing substantial error reductions when the input space is enriched.  This suggests that transformer-based architectures are more adept at capturing the cross-channel dependencies essential for simulating complex hydrological responses. 
Beyond architectural differences, the ablation reveals a critical physical insight regarding the hierarchy of environmental drivers. Across the most responsive models, groundwater consistently emerges as the most informative auxiliary factor, far outweighing the predictive contribution of immediate rainfall or control data.  The exclusion of groundwater (denoted as ``w/o G'' or ``w/o GR'') leads to a significantly higher degradation in MSE and MAE compared to the removal of other variables, such as ``w/o R'' or ``w/o C''.  Notably, for iTransformer and TimesNet, providing groundwater as the sole auxiliary input obtains the highest performance gains among all ablation settings.  This suggests that subsurface water storage serves as a dominant latent state that governs surface water dynamics during compound events, effectively acting as a long-term memory component of the hydrological cycle.

\begin{figure*}[t] 
    \centering    
    \scalebox{0.90}{
    \begin{subfigure}[h]{0.3\textwidth}
        \includegraphics[width=0.8\textwidth]{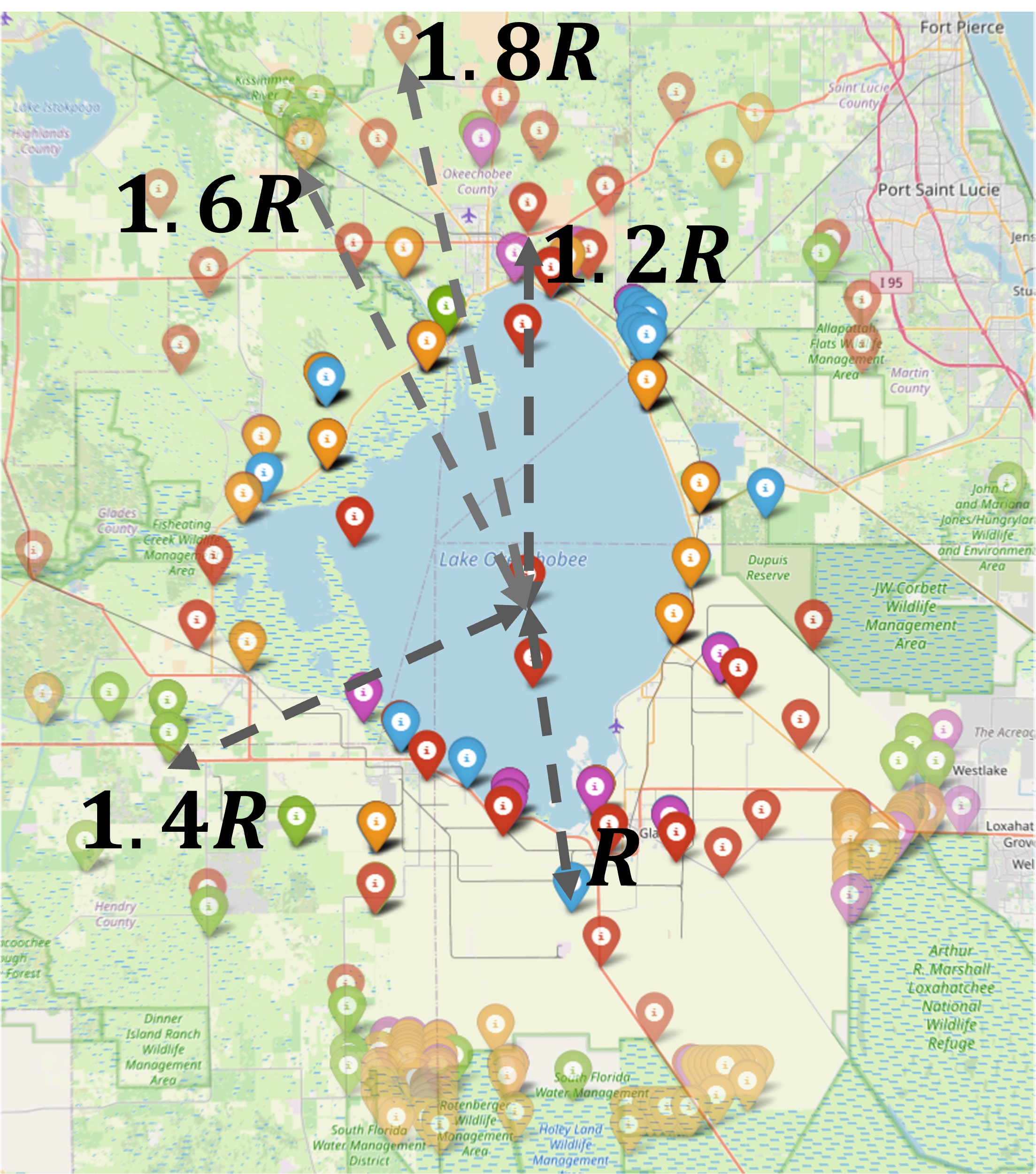}
        \caption{}
        \label{fig:ablation_area}
    \end{subfigure}
    \hspace{-1.0cm}
    \begin{subfigure}[h]{0.65\linewidth}
        \centering
        \begin{subfigure}[b]{0.9\linewidth}
            \includegraphics[width=\textwidth,trim=30 0 20 0, clip]{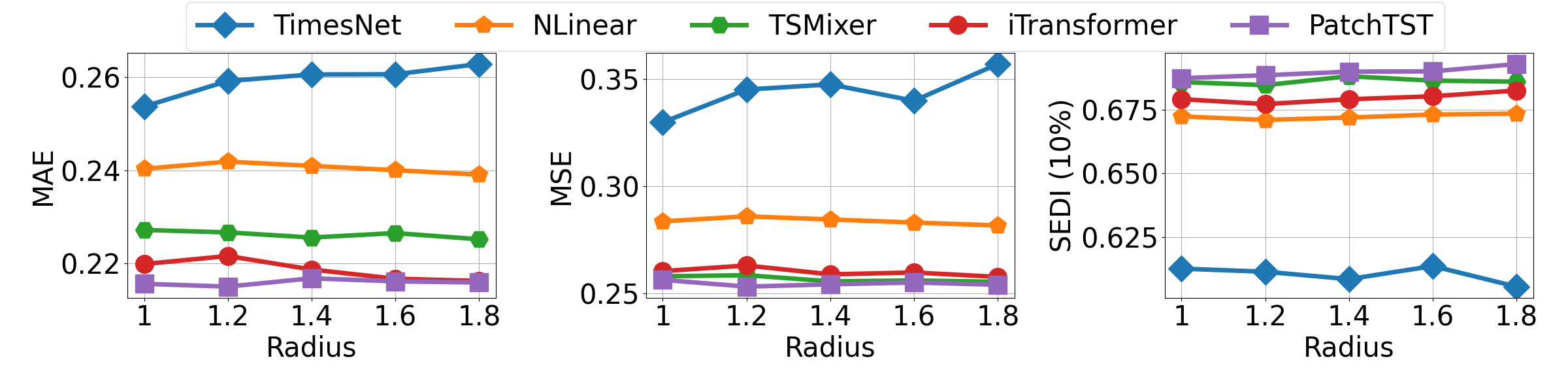}
            \vspace{-0.8cm} 
            \caption{} 
            \label{fig:spatial_ablation}
        \end{subfigure}\\
        \begin{subfigure}[b]{0.9\linewidth}
            \centering
            \includegraphics[width=\textwidth,trim=30 0 20 0, clip]{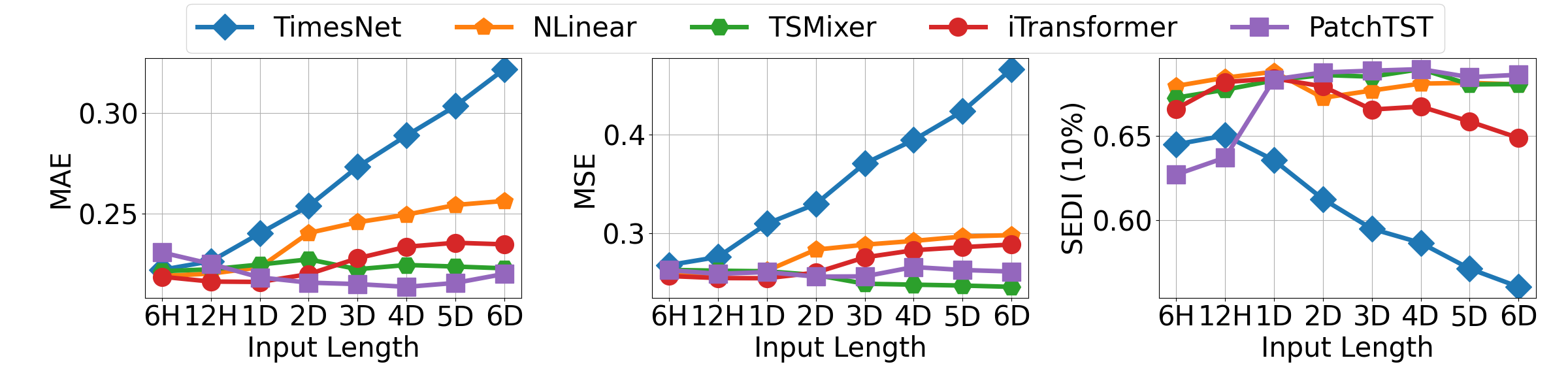}
            \vspace{-0.8cm} 
            \caption{} 
            \label{fig:temporal_ablation}
        \end{subfigure}
    \end{subfigure}
    }
    \vspace{-0.4cm} 
    \caption{(a).The different interest areas with a radius for the ablation study on spatial information. (b). Study on spatial information. (c). Study on temporal information.  }
    \label{fig:combined_ablation}
    \vspace{-0.4cm} 
\end{figure*}

\noindent \textbf{Q4: Is temporal context or spatial context more critical for accurate forecasting?}  

To characterize the boundary of predictability in compound flooding, we investigated the relative importance of spatial coverage and temporal receptivity across different modeling paradigms. 
By defining an anchor area of radius $R$ as the forecasting target and incrementally expanding the interest area using scale factors from 1.0 to 1.8 (as illustrated in Figure~\ref{fig:ablation_area}), we quantified the impact of regional spatial information.  As shown in Figure~\ref{fig:spatial_ablation}, models such as iTransformer, PatchTST, and TSMixer exhibit consistent error reductions as more surrounding monitoring stations are integrated. This performance gain underscores the value of capturing upstream-downstream hydraulic correlations and regional meteorological trends. However, these spatial benefits tend to saturate as the radius increases, suggesting that the marginal utility of geographic data diminishes once the immediate hydrological catchment or local forcing zone is fully represented.

In parallel, we examined the influence of temporal context by varying the input look-back window from 6 hours to 6 days. The results in Figure~\ref{fig:temporal_ablation} reveal that temporal extension generally brings in more substantial performance improvements than spatial expansion, particularly for PatchTST and TSMixer, which achieved significant MSE reductions with longer histories. This dominance of the temporal dimension suggests that the sequential evolution of water stages contains more critical predictive information than instantaneous spatial snapshots, reflecting cumulative rainfall effects and tidal phase transitions. We also observed a performance inflection point for the iTransformer, where accuracy began to degrade beyond a one-day input window.  
This phenomenon reflects a trade-off between the depth of the temporal context and the data density required to resolve long-range dependencies without overfitting. 

In summary, our analysis indicates that while spatial information provides necessary geographical constraints, the temporal context is the primary driver of accuracy, though its benefits are ultimately constrained by the complexity of the underlying physical patterns and the limits of the available training data.

\begin{figure}[t]
    \centering
    \scalebox{0.95}{
    \begin{subfigure}[h]{0.155\textwidth}
        \includegraphics[width=\textwidth]{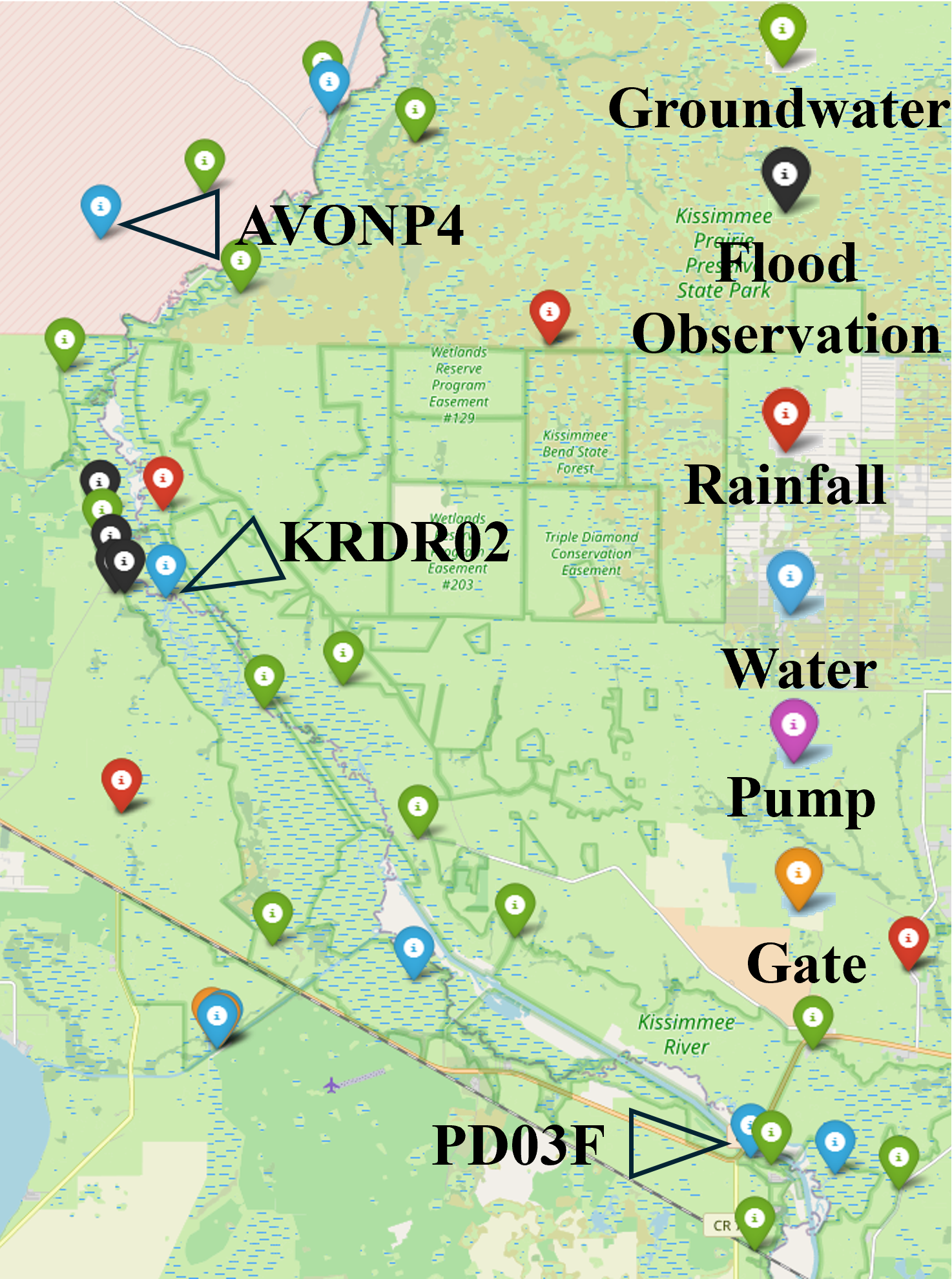}
        \caption{}
        \label{fig:case_flood_area}
    \end{subfigure}
    \begin{subfigure}[h]{0.3\textwidth}
        \includegraphics[width=\textwidth, trim=20 0 15 0, clip]{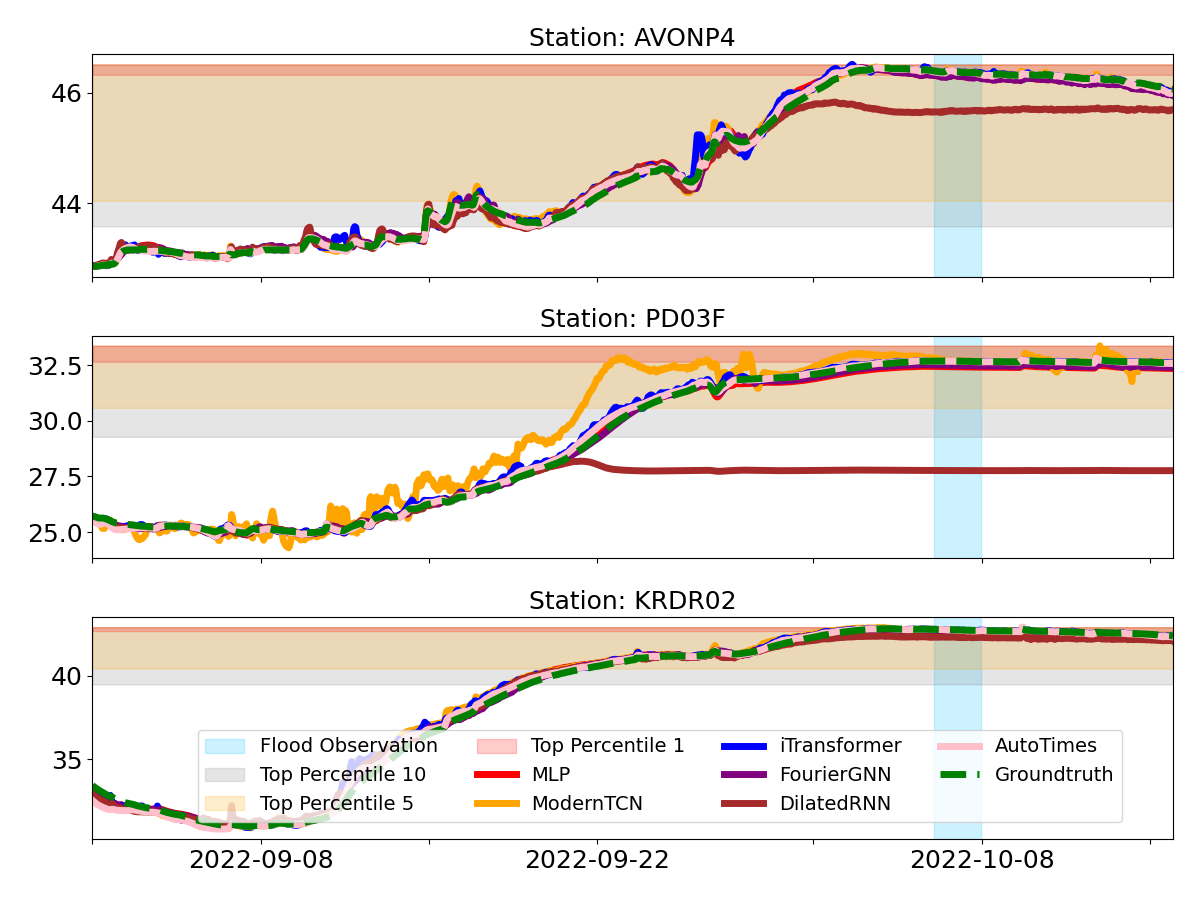} 
    \vspace{-0.6cm}
        \caption{}
        \label{fig:case_flood_results}
    \end{subfigure}
    }
    \vspace{-0.4cm}
    \caption{(a) The interest area of the case study. (b) The comparison among different methods. We mark the flood observation on the x-axis and the top percentile 10/5/1 with colors.}
    \label{fig:case_flood}
    \vspace{-0.65cm}
\end{figure}

\subsection{Compound Flooding Event Study}

To complement the statistical evaluation, we provide an event-level case study on Hurricane Ian, which struck Florida in September 2022 and triggered widespread compound flooding. This event provides a natural stress test for AI models, as it involves intense rainfall, elevated groundwater levels, and rapid surface water response under human regulation. Evaluating model behavior during such an extreme, non-stationary period allows us to assess whether learned dynamics remain physically plausible beyond average conditions.

We focus on the middle reach of the Kissimmee River, where multiple water monitoring stations are located in close proximity to documented flood zones (Figure~\ref{fig:case_flood_area}). This region provides a representative setting for evaluating compound flooding dynamics, with high-quality observations available for key driving factors, including surface water level, groundwater level, rainfall, and human management activities. All models are trained using data before the event and tasked with forecasting water stage trajectories over the subsequent 24-hour period, enabling a direct comparison of model behavior under extreme, non-stationary conditions.

\noindent \textbf{Q5. How do different AI models respond to rapidly evolving compound flood dynamics under extreme conditions?} Figure~\ref{fig:case_flood_results} summarizes the 24-hour water level forecasting results at three representative monitoring stations along the Kissimmee River. Across all stations, most methods successfully capture the overall rising trend of water levels driven by sustained rainfall. However, clear differences emerge in their ability to track the timing and magnitude of flood peaks.  
ModernTCN and DilatedRNN exhibit delayed responses and reduced peak predictions, suggesting limited capacity to adapt to unexpected regime shifts.  In contrast, transformer-based models demonstrate improved responsiveness to rapid water level increases, although iTransformer shows noticeable overestimation and vibrant behavior at certain stations, indicating potential sensitivity to noise amplification under highly non-stationary inputs.  These qualitative observations are consistent with the quantitative evaluation reported in Table~\ref{tab:avg_part_all_msemae_new}, reinforcing that accurate compound flood forecasting requires not only trend-following capability but also robust temporal alignment and stability under extreme hydrological forcing.

\section{Conclusion}
We investigate the use of AI-based forecasting models for compound floods forecasting. By combining standard accuracy metrics with SEDI, we provide a comprehensive assessment of model performance. Our results reveal that while AI models can capture the overall flooding trend, their reliability in predicting extreme water level responses varies substantially.  The case study demonstrates that model failures often emerge during rapid regime shifts driven by the interaction of multiple factors.  These findings reveal the shortcomings of existing data-driven AI methods and explore the driving factors behind compound flooding events. The future work will focus on developing more effective data-driven models capable of integrating multiple driving factors from spatiotemporal information, as well as constructing interpretable frameworks to support the prevention and management of compound floods. In particular, such frameworks should help elucidate how models balance and weigh competing drivers, such as storm surge and peak river discharge, in shaping compound flood dynamics.

\section*{GenAI Disclosure}  
In this paper, we leverage LLMs, including ChatGPT and Gemini, to refine sentence-level writing.

\section*{Acknowledgement}
This project was partially supported by NSF grants IIS-2529283, IIS-2331908, and ECCS-2242700. The views
and conclusions contained in this paper are those of the authors and should not be interpreted as
representing any funding agencies.

\bibliographystyle{ACM-Reference-Format}
\bibliography{sample-base}

\appendix
\section{Time Split of {\dataset}}
\label{appendix:detailed}

We provide the detailed information about the number of monitor stations in each split in Table ~\ref{tab:splits}. In each split, the number of water monitor stations is more than other kinds of features, which means the water level feature is the main information, and others are additional parts. 
\begin{table}[!h]
\centering
\vspace{-0.3cm}
\caption{The summary of the all splits in {\dataset}} 
\vspace{-0.3cm}
\label{tab:splits}
\scalebox{0.78}{
    \begin{tabular}{lcccccccc}
        \toprule
     Splits    &  Time Span &  Interval  &  Water & Groundwater &  Rainfall & Pump    & Gate  \\
     \midrule
     S0        &  1985-1990 & 1 Hour  &  159 & 40  & 143 & 17 & 82 \\ 
     \rowcolor{gray!20} S1        &  1990-1995 & 1 Hour  &  227 & 36  & 139 & 18 & 104\\ 
     S2        &  1995-2000 & 1 Hour  &  332 & 44  & 170 & 26 & 94\\ 
     \rowcolor{gray!20} S3        &  2000-2005 & 1 Hour  &  402 & 178 & 227 & 31 & 107\\ 
     S4        &  2005-2010 & 1 Hour  &  518 & 296 & 254 & 48 & 172\\ 
     \rowcolor{gray!20} S5        &  2010-2015 & 1 Hour  &  585 & 333 & 216 & 65 & 256\\ 
     S6        &  2015-2020 & 1 Hour  &  670 & 317 & 186 & 85 & 300\\ 
     \rowcolor{gray!20} S7        &  2020-2024 & 1 Hour  &  716 & 352 & 194 & 89 & 329\\ 
    \bottomrule
    \end{tabular}
    }   
\vspace{-0.5cm}
\end{table}

\section{Data Preprocessing}

\textbf{Normalization}. To make the data have a zero mean and unit variance, we follow ~\citep{franceschi2019unsupervised,zhou2021informer,yue2022ts2vec} using z-score to normalize the time series data. For the time series, whose variance is less than 1E-4, we regard its variance as one to avoid the variable overflow(inf or NaN). For forecasting tasks, all the report metrics are based on the normalized data. 

\textbf{Timestamp Features}. Because of some methods, such as NLinear, we do not consider extracting the timestamp feature as part of the input. For those methods that are capable of handling the timestamp feature, we ignore this part. In our code, we also provide timestamp features extracted by following  ~\citep{zhou2021informer,yue2022ts2vec} for further works.

\section{Methods}
\label{appendix:method}
\textbf{MLP}. We implement a classical three-layer MLP with ReLU as the activation function. The input layer dimension is determined by the input sequence length, and the output layer dimension corresponds to the forecast horizon. The hidden layer dimension is 64. For training, the learning rate is $1 \times 10^{-4}$, weight decay is $1 \times 10^{-6}$, and the batch size is 64. The model is trained for 15 epochs. \\
\textbf{NLinear}~\citep{zeng2023transformers}. This is a simple linear model that treats each time series independently, modeling future values using a linear transformation of the most recent input values. Implementation is based on the NeuralForecast library\footnote{\url{https://github.com/Nixtla/neuralforecast}}. The learning rate is $1 \times 10^{-4}$, weight decay is $1 \times 10^{-6}$, batch size is 64, and training is performed for 50 epochs. \\
\textbf{TSMixer}~\citep{chen2023tsmixer}. Inspired by MLP-Mixer models from vision tasks, TSMixer is a neural network architecture for time series forecasting. It alternately applies MLPs along the time and feature axes, learning dependencies across both dimensions without requiring attention mechanisms or complex sequence modeling. Implementation follows the NeuralForecast default settings. The architecture includes 2 mixing layers, and the second feed-forward layer has 64 units. The learning rate is $1 \times 10^{-4}$, batch size is 32, and the model is trained for 10 epochs.\\
\textbf{TCN}~\citep{bai2018empirical}. It incorporates causal convolutions, ensuring predictions depend only on current and past inputs, thus preserving temporal order. They also utilize dilated convolutions to efficiently capture long-range dependencies by expanding the receptive field without significantly increasing layers. Our implementation follows the official code\footnote{\url{https://github.com/locuslab/TCN/tree/master}} using the popular channel-wise setting. It employs a three-layer backbone with a kernel size of 3 and a fixed dilation of 1. The learning rate is $1 \times 10^{-3}$, weight decay is $1 \times 10^{-7}$, batch size is 256, and training is performed for 50 epochs.\\
\textbf{ModernTCN}~\citep{luo2024moderntcn}. ModernTCN introduces a streamlined, fully convolutional architecture that aims to simplify design while enhancing performance. It incorporates components like depth-wise separable convolutions and Gated Linear Units (GLUs) to efficiently capture local and long-range temporal dependencies. Implementation follows the long-term forecasting settings from the source code\footnote{\url{https://github.com/luodhhh/ModernTCN}}, using the Weather dataset hyperparameters as defaults. The learning rate is $1 \times 10^{-4}$, batch size is 256, and the model is trained for 100 epochs.\\
\textbf{TimesNet}~\citep{wu2023timesnet}. TimesNet models temporal variations in a two-dimensional space by reshaping time series data into a pseudo-image format and applying 2D convolutional techniques. This enables it to capture both short-term dynamics and long-term dependencies. Implementation uses the Time-Series-Library\footnote{\url{https://github.com/thuml/Time-Series-Library}}, with default hyperparameters from the long-term forecasting setting for the Weather dataset. The learning rate is $1 \times 10^{-4}$, batch size is 32, and training is performed for 10 epochs.\\
\textbf{LSTM}~\citep{hochreiter1997long}. Our Long Short-Term Memory (LSTM) implementation is a two-layer model with a hidden dimension of 32. The learning rate is $1 \times 10^{-3}$, weight decay is $1 \times 10^{-6}$, batch size is 64, and the model is trained for 50 epochs.\\
\textbf{DeepAR}~\citep{salinas2020deepar}. DeepAR is a global model trained on multiple related time series, which aids generalization, especially for series with limited history. It employs an RNN architecture to predict future values by modeling the conditional distribution of the next value given past observations. Implementation is based on the NeuralForecast library. The learning rate is $1 \times 10^{-3}$, batch size is 64, and training is performed for 20 epochs.\\
\textbf{DilatedRNN}~\citep{chang2017dilated}. Dilated RNNs exponentially expand their receptive field by stacking layers with different dilation factors, allowing efficient capture of short- and long-range patterns without a drastic increase in parameters. This makes them suitable for forecasting tasks with wide-ranging temporal dependencies. Implementation is based on the NeuralForecast library. The learning rate is $1 \times 10^{-3}$, batch size is 64, and training is performed for 40 epochs.\\
\textbf{GCN}~\citep{kipf2017semisupervised}. The GCN architecture consists of two layers with a hidden dimension of 32. The graph topology is derived from location information using Delaunay triangulation\footnote{\url{https://docs.scipy.org/doc/scipy/reference/generated/scipy.spatial.Delaunay.html}}. The learning rate is $1 \times 10^{-4}$, weight decay is $1 \times 10^{-5}$, batch size is 32, and the model is trained for 50 epochs.\\
\textbf{FourierGNN}~\citep{FourierGNN}. FourierGNN leverages graph neural networks and Fourier transforms to capture temporal and inter-variable dependencies. Time series variables are treated as graph nodes, with edges representing their relationships. Fourier transforms project data into the frequency domain to model periodic and long-range dependencies. Implementation follows the source code\footnote{\url{https://github.com/aikunyi/FourierGNN}}. The learning rate is $1 \times 10^{-5}$, batch size is 32, and training is performed for 100 epochs.\\
\textbf{StemGNN}~\citep{stemgnn}. StemGNN is designed to capture both temporal (via temporal convolutions) and spatial (via spectral graph convolutions) dependencies in time-series data, learning smooth representations over the graph structure and dynamic patterns. Implementation follows the source code\footnote{\url{https://github.com/microsoft/StemGNN}}. The learning rate is $1 \times 10^{-4}$, batch size is 32, and the model is trained for 50 epochs. \\
\textbf{iTransformer}~\citep{liu2024itransformer}. The iTransformer uses an encoder-decoder structure where the encoder processes the sequence in reverse order. This allows the decoder to predict future values based on this processed representation, enhancing focus on relevant temporal sequences while mitigating the computational cost of traditional transformers. Implementation is based on the NeuralForecast library. The learning rate is $1 \times 10^{-4}$, batch size is 32, and training is performed for 10 epochs.\\
\textbf{PatchTST}~\citep{nie2023a}. PatchTST is a Transformer-based architecture employing patching and channel independence. Time series are divided into patches, which are transformed into tokens and processed by a transformer model to capture local and global dependencies via self-attention. This is particularly useful for long-term forecasting. Implementation is based on the NeuralForecast library. The learning rate is $1 \times 10^{-4}$, batch size is 128, and training is performed for 100 epochs.\\
\textbf{GPT4TS}~\citep{onefitsall}. GPT4TS treats time series as a language, leveraging pretrained language models (LLMs) to learn temporal patterns. Time series data is tokenized for LLM processing, enabling zero-shot or few-shot generalization. Implementation follows the source code\footnote{\url{https://github.com/DAMO-DI-ML/NeurIPS2023-One-Fits-All}}, using long-term forecasting settings for the Weather dataset as default hyperparameters. The default language model is GPT-2~\citep{radford2019language}. The learning rate is $1 \times 10^{-4}$, batch size is 64, and training is performed for 10 epochs.\\
\textbf{AutoTimes}~\citep{liu2024autotimes}. AutoTimes projects time series segments into the embedding space of language tokens, leveraging the autoregressive capabilities of LLMs for forecasting. By training the model to predict subsequent time series segments given preceding ones, AutoTimes generates multi-step forecasts. Implementation follows the source code\footnote{\url{https://github.com/thuml/AutoTimes/tree/main}}, with GPT-2~\citep{radford2019language} as the language model. The learning rate is $5 \times 10^{-4}$, batch size is 64, and training is performed for 10 epochs.\\
For all benchmark experiments, AdamW~\citep{loshchilov2018decoupled} is used as the optimizer, and the loss function is Mean Squared Error (MSE).

\begin{table}
    \centering
    \captionof{table}{Average results of basic methods on the whole dataset. The best and second results are shown in bold font and underlined, respectively. The best performances are in \textcolor{cyan}{\textbf{bold}}, and the second-best method is \textcolor{orange}{\underline{underlined}}. } 
    \vspace{-0.3cm}
    \label{tab:avg_all}
    \scalebox{1.0}{
        \begin{tabular}{l|cccc}
            \toprule
           Metric & \multicolumn{1}{c}{MLP} 
           & \multicolumn{1}{c}{LSTM} 
           & \multicolumn{1}{c}{TCN} 
           & \multicolumn{1}{c}{GCN} \\
        \midrule
         $\downarrow$MAE      & \textcolor{cyan}{\textbf{0.2328}}	   & 0.3302	   & 0.3491    & \textcolor{orange}{\underline{0.3053}}	 \\
        \rowcolor{gray!20} $\downarrow$MSE      & \textcolor{cyan}{\textbf{0.3902}}      & 2.2772   & \textcolor{orange}{\underline{0.6807}}	   & 1.0673	 \\
        $\uparrow$SEDI(10\%)  & \textcolor{cyan}{\textbf{0.4670}}	    & 0.3367	& 0.1689	& \textcolor{orange}{\underline{0.4136}}	\\
        \rowcolor{gray!20} $\uparrow$SEDI(5\%)   & \textcolor{cyan}{\textbf{0.3549}}	    & 0.2121	& 0.1221	& \textcolor{orange}{\underline{0.3118}}	\\
        $\uparrow$SEDI(1\%)   & \textcolor{cyan}{\textbf{0.1718}}     & 0.0545	& 0.0571	& \textcolor{orange}{\underline{0.1477}}	\\
         \bottomrule
        \end{tabular}
    }
    \vspace{-0.4cm}
\end{table}

\section{Platform}
All experiments are conducted on two Linux machines, one with 8 NVIDIA A100 GPUs, each with 40GB of memory, and another with 4 RTX 4090 GPUs. We used Python 3.12.9 and Pytorch 2.6.0 to construct our project.

\section{Results on All Stations}
\label{appendix:extend_res}
For the entire dataset, we report the results of the basic foundational techniques. We provide the results in Table~\ref{tab:avg_all}. As the results show, MLP achieves the best performance, and GCN is the second-best, consistent with previous results in Section~\ref{sec:discovery}.

\end{document}